\documentclass[10pt,twocolumn,letterpaper]{article}

\usepackage{iccv}

\usepackage{graphicx}
\usepackage{amsmath}
\usepackage{amssymb}
\usepackage{booktabs}
\usepackage{listings}
\usepackage{enumitem}
\usepackage{verbatim}
\usepackage{pifont}
\usepackage{amssymb}
\usepackage{dblfloatfix}
\usepackage{footmisc}
\usepackage{graphicx}
\usepackage{booktabs}
\usepackage{float}
\usepackage{dblfloatfix}
\usepackage{xcolor}
\usepackage{multirow}
\usepackage{graphicx}
\usepackage{subcaption}
\usepackage{float}
\usepackage{colortbl}

\usepackage{pifont}
\newcommand{\mycc}{\cellcolor{lightgray!30}}

\usepackage[pagebackref,breaklinks,colorlinks,bookmarks=false]{hyperref}

\usepackage[capitalize]{cleveref}
\crefname{section}{Sec.}{Secs.}
\Crefname{section}{Section}{Sections}
\Crefname{table}{Table}{Tables}
\crefname{table}{Tab.}{Tabs.}

\definecolor{Gray}{gray}{0.9}
\usepackage{amsmath}

\definecolor{gray9}{gray}{.9}
\definecolor{gray95}{gray}{.95}
\definecolor{gray8}{gray}{.8}
\definecolor{gray85}{gray}{.85}

\usepackage{array}
\usepackage{amsmath}
\usepackage{amssymb}
\usepackage{mathtools}
\usepackage{amsthm}
\usepackage{bm}

\newcommand{\roberta}{$\text{RoBERTa}_\text{Large}$}

\newcommand\Bh{\bm{h}}

\newcommand\Bv{\bm{v}}

\newcommand\Bx{\bm{x}}

\newcommand\BA{\bm{A}}
\newcommand\BB{\bm{B}}
\newcommand\BC{\bm{C}}

\newcommand\BF{\bm{F}}

\newcommand\BI{\bm{I}}

\newcommand\BP{\bm{P}}
\newcommand\BQ{\bm{Q}}
\newcommand\BR{\bm{R}}
\newcommand\BS{\bm{S}}

\newcommand\BU{\bm{U}}
\newcommand\BV{\bm{V}}
\newcommand\BW{\bm{W}}
\newcommand\BX{\bm{X}}
\newcommand\BY{\bm{Y}}

 \newcommand{\dR}{\mathbb{R}}

\makeatletter
\newcommand{\dlmf}[1]{%
\citep[%
  \def\nextitem{\def\nextitem{, }}%
  \@for \el:=#1\do{\nextitem\href{http://dlmf.nist.gov/\el}{(\el)}}%
]{olver_nist_2010}%
}
\makeatother

\definecolor{best}{RGB}{0, 204, 0}%
\definecolor{second}{RGB}{178, 255, 102}

\makeatletter
\newcommand{\settitle}{\@maketitle}
\makeatother

\definecolor{deepgreen}{rgb}{0.0, 0.5, 0.0}

\def\paperID{5966}
\def\confName{ICCV}
\def\confYear{2025}

\begin{document}

\title{ConsNoTrainLoRA: Data-driven Weight Initialization of Low-rank Adapters using Constraints}  





\author{
Debasmit Das \thanks{These authors contributed equally to this work.}
\quad
Hyoungwoo Park \footnotemark[1]
\quad
Munawar Hayat
\quad
Seokeon Choi
\quad
Sungrack Yun
\quad
Fatih Porikli\\
{Qualcomm AI Research \thanks{\fontsize{7.5}{11}\selectfont Qualcomm AI Research is an initiative of Qualcomm Technologies, Inc.} }\\
}

\maketitle
\thispagestyle{empty}

\begin{abstract}


Foundation models are pre-trained on large-scale datasets and subsequently fine-tuned on small-scale datasets using parameter-efficient fine-tuning (PEFT) techniques like low-rank adapters (LoRA). In most previous works, LoRA weight matrices are randomly initialized with a fixed rank across all attachment points.
In this paper, we \textbf{improve convergence} and \textbf{final performance} of LoRA fine-tuning, using our proposed data-driven weight initialization method, \textbf{ConsNoTrainLoRA} (CNTLoRA).
We express LoRA initialization as a domain shift problem where we use multiple constraints relating the pre-training and fine-tuning activations. By reformulating these constraints, we obtain a \textbf{closed-form estimate} of LoRA weights that \textbf{depends} on \textbf{pre-training weights} and \textbf{fine-tuning activation vectors} and hence requires \textbf{no training} during initialization. This weight estimate is decomposed to initialize the up and down matrices with proposed flexibility of variable ranks. With the proposed initialization method, we fine-tune on downstream tasks such as \textbf{image generation, image classification and image understanding}. Both quantitative and qualitative results demonstrate that CNTLoRA outperforms standard and data-driven weight initialization methods. Extensive analyses and ablations further elucidate the design choices of our framework, providing an optimal recipe for faster convergence and enhanced performance.

\end{abstract}


\begin{figure}[t]
    \centering
    \includegraphics[width=1.0\linewidth]{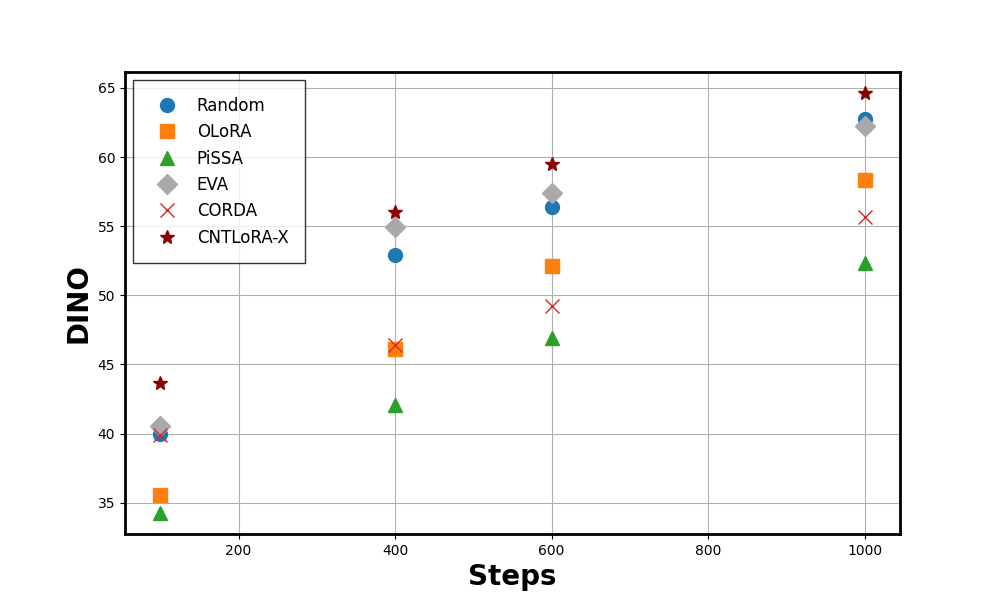}
    \vspace{- 1.5 em}
    \caption{The DINO score plot of different initialization methods with increasing steps of training on the Dreambooth~\cite{ruiz2023dreambooth} dataset. 
    }
    \label{fig:img_conv_15_dino}
    \vspace{- 1.25 em}
\end{figure}

\vspace{-1.00 em}
\section{Introduction}


Foundation models (FMs)~\cite{bommasani2021opportunities} generally undergo an initial phase of large-scale pre-training, followed by task-specific fine-tuning to address domain shift~\cite{mao2024survey} between pre-training and task-specific data.   
This two-step approach has led to advances across multiple applications in natural language processing~\cite{achiam2023gpt, touvron2023llama, team2024gemini}, computer vision~\cite{dehghani2023scaling, oquab2023dinov2} and others~\cite{brohan2023rt, brohan2022rt}. 
However, as the size scales up vastly (or significantly), fine-tuning all parameters of the FM would require much more computational complexity and memory.

Parameter-efficient fine-tuning (PEFT) methods provide a promising solution \eg Low-Rank Adaptation (LoRA) \cite{hu2021lora} only adds a trainable adapter with fewer parameters while keeping the pre-trained model frozen. 
To expedite the convergence of PEFT models, recent works~\cite{meng2024pissa, buyukakyuz2024olora,paischer2024one,yangcorda} explore adapter initialization from model weights or data to encapsulate pre-trained knowledge or task information. 
\begin{figure*}[h]
    \centering
    \includegraphics[width=1.0\linewidth]{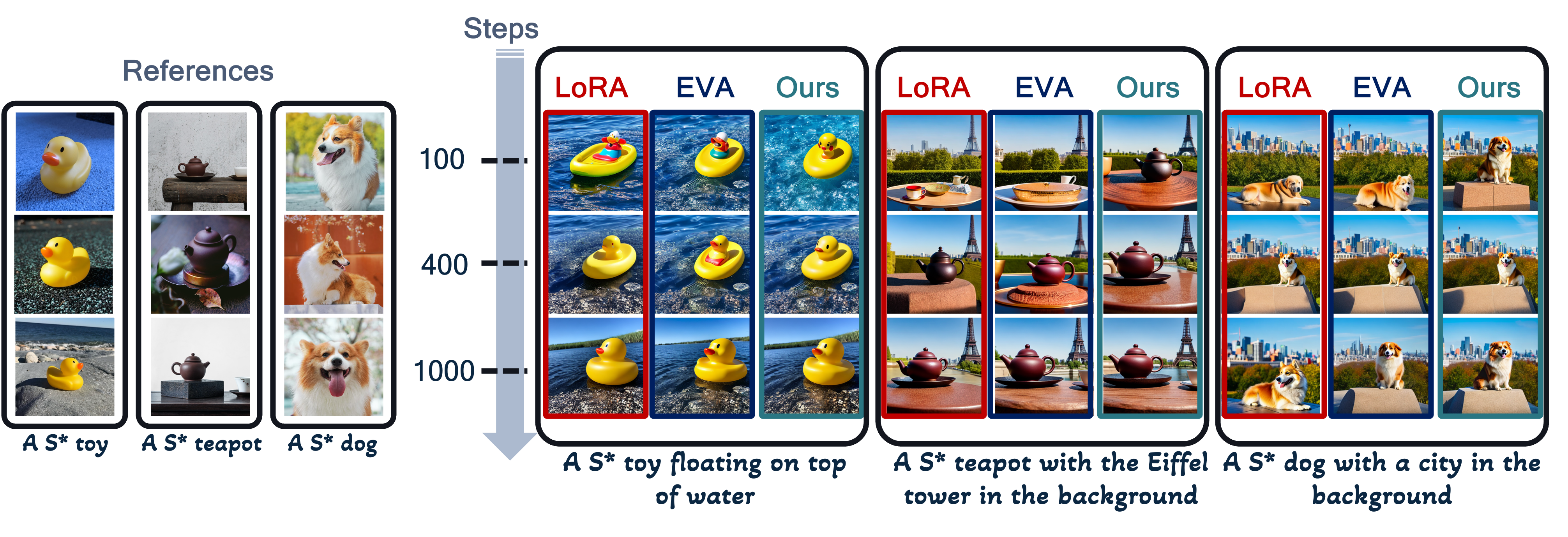} 
    \caption{Generation evolution for different initialization methods on Dreambooth~\cite{ruiz2023dreambooth} dataset. The prompts from left to right are (a) A S* toy floating on top of water. (b) A S* teapot with the Eiffel tower in the background. (c) A S* dog with a city in the background.}
    \label{fig:img_frame}
    \vspace{- 1.25 em}
\end{figure*}
Among the adapter initialization methods, PiSSA~\cite{meng2024pissa} and OLoRA~\cite{buyukakyuz2024olora} are the most popular. PiSSA uses SVD decomposition of pre-trained matrices to initialize LoRA weights, while OLoRA~\cite{buyukakyuz2024olora} uses QR decomposition. Neither method uses fine-tuning data for initialization. EVA \cite{paischer2024one} addresses this by minimizing the reconstruction error of LoRA layer input activations and initializing down matrices from SVD of input activations. However, up matrices are initialized with zeros, making it difficult to balance pre-training knowledge with fine-tuning data. 
To encapsulate both previous and novel knowledge, CORDA~\cite{yangcorda} uses low-rank SVD on the interaction between LoRA weight matrices and the covariance of input activations to initialize up and down matrices. While effective for language tasks, its application to visual tasks remains untested, and the selection of interaction is rather unclear. 
This paper presents a systematic, constraint-based framework for LoRA weight initialization, which promotes faster convergence, improves performance, and also provides a unified  paradigm for existing methods.

We motivate our proposed LoRA initialization method through the lens of domain shift~\cite{tzeng2014deep, tzeng2017adversarial, ganin2016domain, sun2016deep}, which LoRA is proposed to address and resolve the problem. Domain shift in each LoRA layer can be expressed through differences between the pre-training and fine-tuning activations. In the LoRA layer, the input activations are processed by a linear weight matrix to produce output activations.
Accordingly, we could consider four variables of interest in the LoRA layer: (a) pre-training (source) input activations (b) pre-trained (source) weights (c) fine-tuning (target) input activations (d) fine-tuned (target) weights. We propose domain shift constraints to relate all these four variables of interest with the aim to obtain an initial estimate of the target weights. However, source input activations are not available during fine-tuning. Hence, we propose using \emph{rough assumptions} or constraints about the source input activations. The domain shift constraint can be applied between the source and target output activations. We can gather these two constraints to form a constraint set and obtain the target weights. Depending on how to select constraints, we have multiple variants of constraint sets. In this paper, we consider three types of constraint sets, which we refer to 
as \emph{modes} for initialization, see Section~\ref{sec:lora_init_const} for details. 

We use constraint sets to obtain target weight of each LoRA layer. The target weight can then be subtracted from source weight to obtain LoRA weights, which can be further decomposed to obtain up and down matrices of fixed rank. However, having fixed-rank adapters across all attachment points in a network is sub-optimal. Hence, we propose variable adapter structure (VAS) to allocate variable ranks across different attachment points. The relative ranking can be obtained from the decomposition technique, which is generally SVD. We propose using relative variance of singular values in SVD decomposition to select adapter ranks. More details of the procedure are in Section~\ref{sec:special-cases}.

\begin{figure*}[t]
  \centering
  \includegraphics[width=1.0\linewidth]{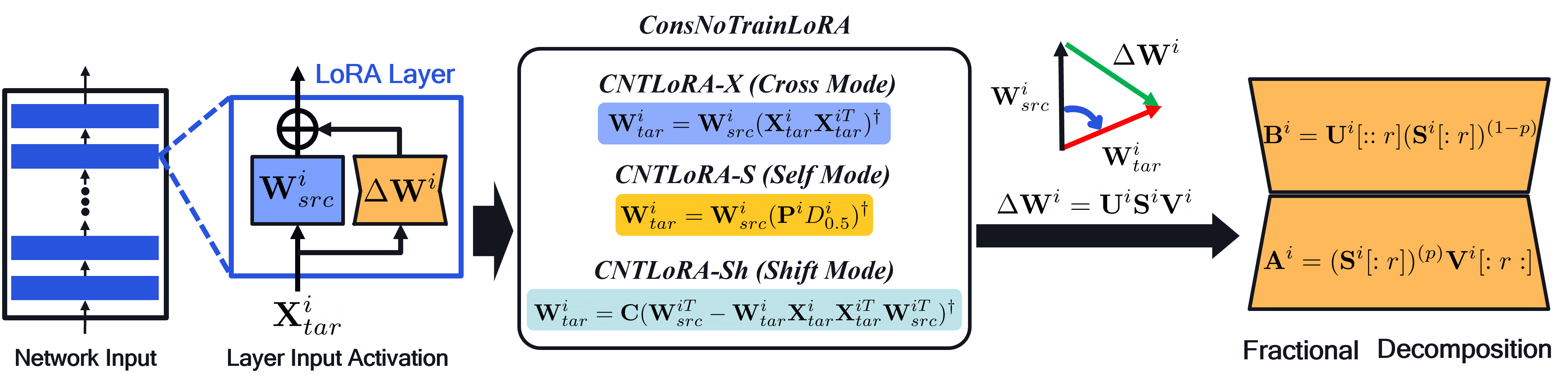}
   \vspace{- 2.0 em}
   \caption{Visualization of our proposed framework. The input samples are fed to the network and the activations before the LoRA layer are used for initialization of the LoRA weight matrix $\Delta \mathbf{W}^{i}$, which can then be decomposed to the up and down matrices.}
   \vspace{-1.5 em}
   \label{fig:motive}
\end{figure*}

Our proposed framework is holistic in the sense that it can initialize both the parameters and structure of the adapter. Our 
framework, as depicted in Fig.~\ref{fig:motive}, provides flexible design choices and is more effective than existing data-driven solutions. Since we use \textbf{constrain}ts with \textbf{no} \textbf{train}ing for LoRA initialization, we call our framework as \textbf{ConsNoTrainLoRA} (CNTLoRA).  We demonstrate the effectiveness of our approach through quantitative and qualitative results in Fig.~\ref{fig:img_conv_15_dino} and~\ref{fig:img_frame}. Our approach shows significant gains over competing methods (\eg better DINO scores at step 100 and 1000 in Fig.~\ref{fig:img_conv_15_dino}) and better image fidelity at even 100 iterations (Fig.~\ref{fig:img_frame}). In summary, our contributions are:
\begin{itemize}[noitemsep, nolistsep]
    \item We propose a suite of modes for data-driven initialization of LoRA. Each mode corresponds to an unique constraint set among activations and weights to obtain a better estimate of the fine-tuned weights, enabling faster and better convergence. 
    \item Our approach enables the flexibility of variable ranks across different adapter attachment points using relative significance of singular values during weight decomposition.
    \item We demonstrate the effectiveness of our framework across a variety of discriminative and generative tasks involving both vision and language modalities. We explore how different design choices and variations of our framework can lead to different performance outputs.
\end{itemize}

\section{Related Work}
\label{RelatedWork}

\subsection{Low-rank adapter and its Variants}

Fine-tuning using low-rank adaptation was introduced in~\cite{hu2021lora}, and has garnered significant attention. Multiple follow-up works~\cite{kopiczko2024elora, zi2023delta, babakniya2023slora, dettmers2024qlora, li2023loftq, nikdan2024rosa, liu2024alora, zhang2023adalora, hayou2024lora+, chavan2023one} have been introduced, striving to improve latency, accuracy, and memory efficiency. Some of the popular ones include AdaLoRA~\cite{zhang2023adalora}, DoRA~\cite{liu2024dora}, LoRA-GA~\cite{wang2024lora}, and LoRA-XS~\cite{balazy2024lora}. AdaLoRA focuses on adaptive rank allocation for different attachment points in the architecture. LoRA-GA approximates the gradient of the pre-trained matrix using SVD. LoRA-XS computes SVD of pre-trained weights for LoRA weights and only updates the singular values for better transferability to another model. Additionally, Rank-stabilized LoRA (rsLoRA)~\cite{kalajdzievski2023rank} was proposed to adjust the scaling factor for fine-tuning compute/performance trade-offs. In this paper, we compute the SVD of the LoRA matrix to extract the up and down matrices.



\subsection{Weight Initialization of Low-rank Adapter}

Weight initialization of neural networks is a long-standing topic. The early works of He et al.~\cite{he2015delving} and Glorot \& Bengio~\cite{glorot2010understanding} developed methods to ensure stable training of deep networks by considering activation functions and network depth. For parameter-efficient fine-tuning,~\cite{hu2021lora, liu2022few} considered data-driven initialization by pre-training on a related task or unsupervised pre-training on the fine-tuning task. In~\cite{nikdan2024rosa}, adaptation is done to initialize a sparse matrix. Babakniya et al.~\cite{babakniya2023slora} used SVD on weight matrices after some fine-tuning to initialize the LoRA matrices. Recently, model-driven weight initialization has gained attention, and~\cite{meng2024pissa, buyukakyuz2024olora} use pre-trained weight information followed by decomposition to initialize the LoRA weights. EVA~\cite{paischer2024one} considers initialization of down matrices using singular vectors of activations. 
Recent works~\cite{wang2024lora, yangcorda} also consider data-driven initialization but do not consider adaptive rank allocation. Specifically, they consider the intermediate activations of the LoRA layer and compute their statistics followed by decomposition to obtain the up and down matrices. Our proposed framework is a hybrid technique that considers both model-driven initialization and data-driven initialization and can also seamlessly integrate adaptive rank allocation if required.

\subsection{Efficient fine-tuning of Low-rank adapter}

Another line of work examines improving efficiency of LoRA fine-tuning. This includes memory-efficient techniques like~\cite{kopiczko2024elora}, which keep LoRA matrices frozen and only update scaling vectors. Other works improve memory and latency, including quantization techniques~\cite{dettmers2022gpt3} integrated with LoRA. Interestingly, there are other variants of LoRA~\cite{dettmers2022gpt3} that can integrate quantization into their framework. Furthermore, customized initialization has also improved the fine-tuning results of quantized models~\cite{nikdan2024rosa, valipour2022dylora,meng2024pissa}. In this paper, we consider improving convergence and final performance of fine-tuning. 
While memory efficiency is not our primary objective, however, our variable adapter rank is capable of achieving this efficiency.

\section{Method}

\subsection{Background on LoRA}
\label{sec:background}


The rationale for using low-rank matrices is the assumption that fine-tuning data has low intrinsic dimensionality and hence loss gradients are also low-rank~\cite{aghajanyan2020intrinsic}. 
The input to the LoRA layer is: $\Bx \in \dR^{d\times1}$ which is passed to the pre-trained weight matrix $\BW \in \dR^{k\times d}$. The LoRA layer introduces new trainable low-rank matrices $\BA$ and $\BB$ such that $\Bh = \BW\Bx + \BB\BA\Bx$, where $\BB \in \dR^{k \times r}$ and $\BA \in \dR^{r \times d}$. The rank $r\ll k$ is a constant that needs to be fixed before fine-tuning. During fine-tuning, $\BW$ is kept constant but $\BA$ and $\BB$ are updated. In the original LoRA paper, it was proposed to have $\BB$ initialized with zeros and $\BA$ initialized randomly. Recent methods strive to initialize $\BB$, $\BA$ from the pre-trained weights $\BW$ and/or $\Bx$. This is generally done to improve convergence and final performance scores. Furthermore,  a constant is used to scale $\BB\BA\Bx$ by $\frac{\alpha}{r}$.

\subsection{ConsNoTrainLoRA}
\label{sec:lora_init_const}
Our proposed initialization method is motivated by the domain shift between source and target activations. Here, source activations are produced from the pre-training data, while target activations are obtained from the fine-tuning data. Consider that the activation inputs for a LoRA layer $i$ from the source domain are denoted as $\BX^{i}_{src} \in \dR^{d\times b}$ and those from the target domain are denoted as $\BX^{i}_{tar} \in \dR^{d\times b}$. We consider the pre-trained weight of a LoRA layer as $\BW^{i}_{src} \in \dR^{k\times d}$. Let the estimate of the initialization be denoted as $\BW^{i}_{tar} \in \dR^{k\times d}$ such that $\BW^{i}_{tar} = \BW^{i}_{src} + \Delta \BW^{i}$. Our goal is to estimate $\BW^{i}_{tar}$ from the variables $\BX^{i}_{src}$, $\BX^{i}_{tar}$, and $\BW^{i}_{src}$. The caveat is that we have another unknown $\BX^{i}_{src}$, which is not available during fine-tuning. Hence, we need to use multiple constraints on these variables so that the estimate of $\BW^{i}_{tar}$ can be obtained. It is important to note that \textbf{these constraints are rough assumptions} just to provide a good initialization. After the initialization, we would still constrain adapter training using standard task objectives. We can represent them as a constraint set such that $\BF_j(\BX^{i}_{src}, \BX^{i}_{tar}, \BW^{i}_{src}, \BW^{i}_{tar}) = 0$ where $j=1, 2, .... N$. Since we have two unknown variables, we set $N=2$. We propose three constraint sets that are aimed to capture different statistical relations between the source and target activations. Now, we proceed to describe the different constraint sets or modes (as we shall call them in the paper) for initialization. 


\noindent \textbf{Cross Mode}\quad In this mode, we mainly consider first order matching between the source and target domain. This is achieved through two constraints: (1) Output activations of the source and target domain are equal. (2) Input activations of the source and target domain are highly correlated. Constraint (1) can be realized as 
\begin{equation}\label{eq:cross1}
\BF_1(\BX^{i}_{src}, \BX^{i}_{tar}, \BW^{i}_{src}, \BW^{i}_{tar}) = \BW^{i}_{src} \BX^{i}_{src} - \BW^{i}_{tar} \BX^{i}_{tar} = \mathbf{0}
\end{equation}
Constraint (2) can be realized as  
\begin{equation}\label{eq:cross2}
\BF_2(\BX^{i}_{src}, \BX^{i}_{tar}, \BW^{i}_{src}, \BW^{i}_{tar}) = \BX^{i}_{src} \BX^{iT}_{tar} - \mathbf{I} = \mathbf{0}
\end{equation} 

Equation~\eqref{eq:cross2} can be re-arranged as $\BX^{i}_{src} \BX^{iT}_{tar} = \mathbf{I}$. Equation~\eqref{eq:cross1} can then be re-arranged as $\BW^{i}_{src} \BX^{i}_{src} = \BW^{i}_{tar} \BX^{i}_{tar}$, which when multiplied by $\BX^{iT}_{tar}$ on both sides, we get
\begin{equation}\label{eq:cross3}
\BW^{i}_{src} = \BW^{i}_{tar} \BX^{i}_{tar} \BX^{iT}_{tar}
\end{equation} 
Re-arranging the equation, we can obtain the least squares pseudo-inverse~\cite{moore1920reciprocal} solution as 
\begin{equation}\label{eq:cross4}
\boxed{\BW^{i}_{tar} = \BW^{i}_{src} (\BX^{i}_{tar} \BX^{iT}_{tar})^{\dagger}} \quad \textbf{CNTLoRA-X}
\end{equation}

\noindent \textbf{Self Mode}\quad 
In this mode, we mainly consider second order matching between the source and target domain.
This is achieved through: (1) Covariances of source and target output activations are equal. (2) Input activations of the source domain are whitened. Constraint (1) can be realized as  

\begin{align}\label{eq:self1}
&\BF_1(\BX^{i}_{src}, \BX^{i}_{tar}, \BW^{i}_{src}, \BW^{i}_{tar}) = \mathbf{0} \\
&\BW^{i}_{src} \BX^{i}_{src} \BX^{iT}_{src} \BW^{iT}_{src} - \BW^{i}_{tar} \BX^{i}_{tar} \BX^{iT}_{tar} \BW^{iT}_{tar} = \mathbf{0}
\end{align}

Constraint (2) can be realized as  
\begin{equation}\label{eq:self2}
\BF_2(\BX^{i}_{src}, \BX^{i}_{tar}, \BW^{i}_{src}, \BW^{i}_{tar}) = \BX^{i}_{src} \BX^{iT}_{src} - \mathbf{I} = \mathbf{0}
\end{equation} 

Equation~\eqref{eq:self2} can be re-arranged as $\BX^{i}_{src} \BX^{iT}_{src} = \mathbf{I}$. Equation~\eqref{eq:self1} can then be re-arranged as $\BW^{i}_{src} \BX^{i}_{src} \BX^{iT}_{src} \BW^{iT}_{src} = \BW^{i}_{tar} \BX^{i}_{tar} \BX^{iT}_{tar} \BW^{iT}_{tar}$, which can be rewritten as
\begin{equation}\label{eq:self3}
\BW^{i}_{src} \BW^{iT}_{src} = \BW^{i}_{tar} \BX^{i}_{tar} \BX^{iT}_{tar} \BW^{iT}_{tar}
\end{equation} 
\eqref{eq:self3} is a quadratic equation, with multiple solutions for $\BW^{i}_{tar}$. However, we consider a specific solution obtained from SVD of $\BX^{i}_{tar} \BX^{iT}_{tar}$. The SVD is obtained as
\begin{equation}\label{eq:self4}
\BX^{i}_{tar} \BX^{iT}_{tar} = \BP^{i} D^{i}\BP^{iT} = \BP^{i} D^{i}_{0.5}D^{i}_{0.5}\BP^{iT}
\end{equation}
Here, $D^{i}$ is a diagonal matrix and $D^{i}_{0.5}$ is its square root. We can substitute $\BX^{i}_{tar} \BX^{iT}_{tar}$ into equation \eqref{eq:self3} as 
\begin{equation}\label{eq:self5}
\BW^{i}_{src} \BW^{iT}_{src} = \BW^{i}_{tar}  \BP^{i} D^{i}_{0.5}D^{iT}_{0.5}\BP^{iT} \BW^{iT}_{tar}
\end{equation}
We can regroup the above equation ~\eqref{eq:self5} into ~\eqref{eq:self6} as 
\begin{equation}\label{eq:self6}
(\BW^{i}_{src}) (\BW^{iT}_{src}) = (\BW^{i}_{tar} \BP^{i} D^{i}_{0.5})(D^{iT}_{0.5}\BP^{iT} \BW^{iT}_{tar})
\end{equation}

Matching the groups and re-arranging the equation, we can obtain the least squares pseudo-inverse solution as 
\begin{equation}\label{eq:self7}
\boxed{\BW^{i}_{tar} = \BW^{i}_{src} (\BP^{i} D^{i}_{0.5})^{\dagger}} \quad \textbf{CNTLoRA-S}
\end{equation}

\noindent \textbf{Shift Mode}\quad
For the third constraint set, we consider modeling the difference in domain shift between source and target features. This is acheived through the following two constraints: (1) Domain Shift with source and target features vary by a constant (2) Source features are whitened. Constraint (1) can be realized as 
\begin{align}\label{eq:shift1}
&\BF_1(\BX^{i}_{src}, \BX^{i}_{tar}, \BW^{i}_{src}, \BW^{i}_{tar}) = \mathbf{0} \\
&\BW^{i}_{tar} \BX^{i}_{src} \BX^{iT}_{src} \BW^{iT}_{src} - \BW^{i}_{tar} \BX^{i}_{tar} \BX^{iT}_{tar} \BW^{iT}_{tar} - \BC = \mathbf{0}
\end{align}
Here, $\BC$ is a hyper-parameter we can vary. Constraint (2) can be realized as  
\begin{equation}\label{eq:shift2}
\BF_2(\BX^{i}_{src}, \BX^{i}_{tar}, \BW^{i}_{src}, \BW^{i}_{tar}) = \BX^{i}_{src} \BX^{iT}_{src} - \mathbf{I} = \mathbf{0}
\end{equation} 

Equation~\eqref{eq:shift2} can be re-arranged as $\BX^{i}_{src} \BX^{iT}_{src} = \mathbf{I}$ and plugged into~\eqref{eq:shift1} to obtain
\begin{equation}\label{eq:shift3}
\BW^{i}_{tar} \BW^{iT}_{src} - \BW^{i}_{tar} \BX^{i}_{tar} \BX^{iT}_{tar} \BW^{iT}_{src} - \BC = \mathbf{0}
\end{equation} 
Re-arranging the equation, we can obtain the least squares pseudo-inverse solution as 
\begin{align}\label{eq:shift4}
\boxed{\BW^{i}_{tar} = \BC(\BW^{iT}_{src} - \BX^{i}_{tar} \BX^{iT}_{tar} \BW^{iT}_{src})^{\dagger}} 
\end{align}
\begin{equation*}
\textbf{CNTLoRA-Sh}
\end{equation*}
By default, we use $\BC$ as identity.

For each of the modes, the estimate can be repeated over all $B$ batches to obtain a robust estimate $\BW^{i}_{est}$ such that 
\begin{equation}\label{eq:avg}
\BW^{i}_{est} = \frac{\BW_0^{i} + \sum_{j=1}^{B}w_j[\BW^{i}_{tar}]_j}{B + 1}
\end{equation} 
where $w_j$ (default value of 1) is the weighing amount of the $j^{th}$ batch.
$\BW_0^{i}$ can be set to the pre-trained weight $\BW^{i}_{src}$ as the default value.

Here, $\BW_{est}^{i}$ can be subtracted from $\BW_{src}^{i}$ to obtain $\Delta\BW_{est}^{i}$, which can then be decomposed with rank $r$ to obtain $\Delta\BW_{est}^{i} = \BU_{est}^{i}[::r] \BS_{est}^{i}[:r] \BV_{est}^{i}[:r:]$, where we can fractionally allocate the singular matrix $\BS_{est}^{i}[:r]$ such that $\BB^{i} = \BU_{est}^{i}[::r] \BS_{est}^{i(1-p)}[:r]$ and $\BA^{i} = \BS_{est}^{ip}[:r] \BV_{est}^{i}[:r:]$. By default, $p=0.5$. Alternatively, $\Delta\BW_{est}^{i}$ can be obtained through QR decomposition as $\Delta\BW_{est}^{i} = \BQ_{est}^{i}[::r] \BR_{est}^{i}[:r:]$ with $\BB^{i} = \BQ_{est}^{i}[::r]$ and $\BA^{i} = \BR_{est}^{i}[:r:]$. 

\subsection{Special cases}
\label{sec:special-cases}

\noindent \textbf{Text to Image Generation}\quad For text-to-image generation, we consider different options for feeding into UNet. During training, a noisy version of the image is fed into the UNet while during inference, the denoising process starts from a noise. For LoRA initialization, we consider inputting a noisy version of the image into the UNet: (a) $x_t = \sqrt{\alpha_t} z_t + \sqrt{1-\alpha_t} \cdot \epsilon$ such that $\epsilon \sim N(0,I)$, where the latent $z_t = E(I_{img})$ and $E$ is the VAE encoder and $I_{img}$ is the input image. Here, $\alpha_t \in (0,1)$ is a variance scheduler. (b) Alternatively, the input can just be the latent embedding $z_t$ without any noise addition. We consider (b) as default.

\begin{figure}[h]
  \centering
  \includegraphics[width=0.9\linewidth]{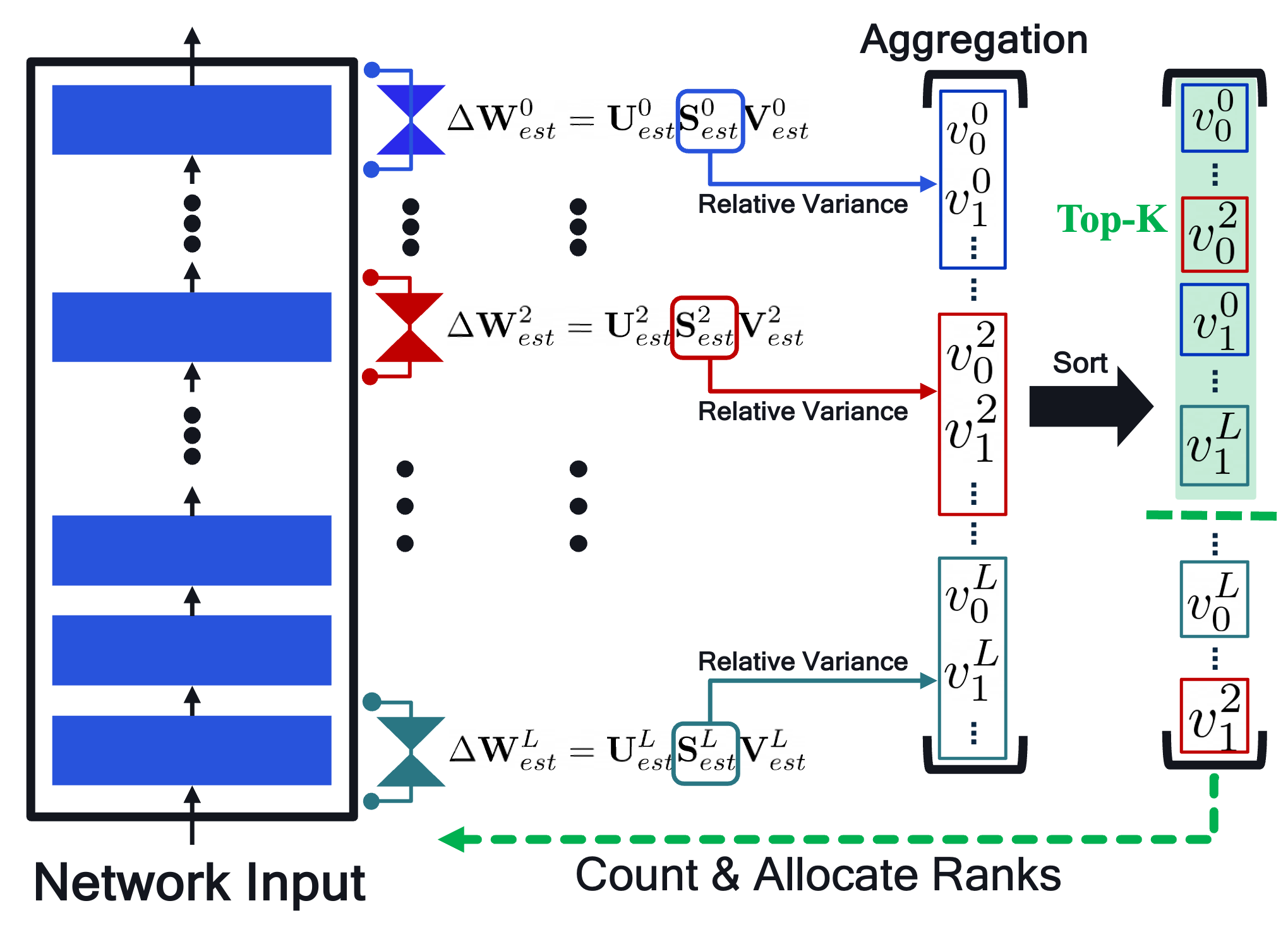}
  \vspace{- 1.0 em}
   \caption{Visualization of how variable ranking  is done to obtain variable adapter structure.}
    \vspace{-0.5 em}
   \label{fig:rankallo}
\end{figure}

\noindent \textbf{Variable Adapter Structure (VAS)} By default, we have fixed and same rank allocation across all attachment points of the foundation model. However, for efficient deployment, there might be restrictions on the adapter sizes. To maintain efficiency, we need adaptive ranking across attachment points. Our framework allows seamless integration for adaptive ranks, especially for the SVD decomposition of $\Delta\BW_{est}^{i} = \BU_{est}^{i} \BS_{est}^{i} \BV_{est}^{i}$ for attachment point $i$. Here, $\BS_{est}^{i}$ is the diagonal singular value matrix, where $s^{i}_{m}$ is the $m^{th}$ singular value corresponding to attachment point $i$. The relative variance $v^{i}_{m}$ of the singular value $s^{i}_{m}$ is given by
\begin{equation}
v^{i}_m \propto s^{i^2}_m/{||\BS_{est}^{i}||_1},
\end{equation}
and the proportion constant is obtained by summing over all the singular values and dividing by the sum. This process of obtaining relative variance $v^{i}_{m}$ is repeated for all attachment points, and we collect them into a long list $\Bv$, which is sorted in descending order, and the top-$K$ values are obtained. The value of $K$ is the rank budget \ie the maximum sum of all the ranks for that particular model. From the $K$ values, we count the contribution from each attachment point $i$ and accordingly assign the count as the rank of the corresponding attachment point $i$. This process of rank allocation is described in Fig.~\ref{fig:rankallo}, and we set the default value of $K$ as the rank times the number of attachment points.

In supplementary, we also visualize VAS and analyze how other LoRA variants can be expressed as constraints and justify convergence of our method using gradient information.

\section{Experiments}
\label{sec:experiments}

\subsection{Experimental Settings}

\noindent \textbf{Image Generation} We evaluate on the {Dreambooth~\cite{ruiz2023dreambooth} dataset}, using DINO and CLIP-I for subject fidelity, and CLIP-T for prompt fidelity. We use {Stable Diffusion v1.5~\cite{rombach2022high}}. During fine-tuning, in addition to denoising loss, we use prior preservation loss weighted at 0.1 with 250 generated class samples. The LoRA~\cite{hu2021lora} is applied to the cross and self-attention layers of UNet and attention layers of text encoder~\cite{radford2021learning}. The default rank and batch size are 128 and 1, respectively. We use AdamW~\cite{loshchilov2017decoupled} optimizer with a learning rate of 1e-4 and fine-tune over 1000 steps.

\noindent \textbf{Image Classification} We consider {VTAB-1K~\cite{zhai2019large}}, which has been the standard benchmark for evaluating PEFT methods. The datasets in this benchmark are used to fine-tune a {DINOv2-g/14} model~\cite{oquab2023dinov2}, which is the same as used in EVA~\cite{paischer2024one}. For fair comparison, our default hyperparameters are those used in EVA~\cite{paischer2024one}.

\noindent \textbf{Image Understanding} We consider two datasets: first, the {Amazon Product Description dataset (APD)}~\cite{amznproddesc} for product marketing. We fine-tune the LoRA with the {GLM-Edge} model~\cite{glm2024chatglm} on this dataset. We further evaluate on {myVLM} dataset~\cite{alaluf2024myvlm}, which consists of concepts for personalized captioning. For fine-tuning, we use {LLAVA-v1.5-7B}~\cite{liu2023visual}. For the APD dataset, we report the SentSim~\cite{reimers-2019-sentence-bert} scores while for the myVLM dataset, we report an additional Recall metric to identify the concept in the generated caption. 

{More details about the implementation, evaluation setup, and additional results are in the supplementary material}

\subsection{Results}
\label{sec:main_results}
\subsubsection{Image generation}
\begin{table}[h]
\caption{Final quantitative results (1000 iterations) of different methods on the Dreambooth dataset using SD 1.5. 
}
\vspace{- 1.5 em}
\label{tab:gen15}
\begin{center}
\resizebox{0.45\textwidth}{!}
{
    \begin{tabular}{l|ccc}
    \toprule
    \textbf{Method} & \textbf{DINO} ($\uparrow$) & \textbf{CLIP-I} ($\uparrow$) & \textbf{CLIP-T} ($\uparrow$)
 \\ 
    \midrule
LoRA & 62.74 & 80.07 & 26.43 \\
OLoRA & 58.36 & 77.56 & 27.23 \\
PISSA & 52.34 & 74.41 & \textbf{27.98} \\
EVA & 62.24 & 79.68 & 25.7 \\
CORDA & 55.65 & 71.44 & 25.6 \\
LoRA-GA & 60.30 & 76.10 & 26.23 \\ \midrule
DoRA & 62.98 & 80.81 & 27.04 \\
RS-LoRA & 63.12 & {81.07} & 26.98 \\
\midrule
\mycc CNTLoRA-X & \mycc {64.63} & \mycc 80.78 & \mycc 25.83 \\
\mycc CNTLoRA-S & \mycc {63.91} & \mycc {81.7} & \mycc 27.07 \\
\mycc CNTLoRA-Sh & \mycc 62.95 & \mycc {80.97} & \mycc \underline{27.87} \\  \midrule
\mycc CNTDoRA-X & \mycc {64.82} & \mycc {81.02} & \mycc {27.67} \\
\mycc CNTRS-LoRA-X & \mycc {64.93} & \mycc {81.61} & \mycc {27.86} \\ \midrule
\mycc CNTLoRA-X + VAS & \mycc \textbf{65.73} & \mycc 81.98 & \mycc 25.63 \\
\mycc CNTLoRA-S + VAS & \mycc \underline{64.94} & \mycc \textbf{82.6} & \mycc 27.17 \\
\mycc CNTLoRA-Sh + VAS & \mycc 63.55 & \mycc \underline{82.03} & \mycc \textbf{27.98} \\
\hline
    \bottomrule
    \end{tabular}}
\end{center}
\vspace{- 1.5 em}
\end{table} 

\noindent \textbf{Quantitative Results} We compare CNTLoRA with EVA, CORDA, LoRA-GA, PISSA, and OLoRA for fine-tuning performance. Table~\ref{tab:gen15} shows that CNTLoRA methods consistently outperform the others. CNTLoRA-X achieves the highest DINO score of 64.63. For CLIP-I, CNTLoRA-S produces score of 81.7, while CNTLoRA-Sh produces score of 80.97. For CLIP-T, CNTLoRA-S and CNTLoRA-Sh produce scores of 27.07 and 27.87, respectively. Integrating CNTLoRA-X with DoRA and RS-LoRA also improves performance. For DINO, CNTDoRA-X produces score of 64.82, and CNTRS-LoRA-X produces a score of 64.93. For CLIP-I, CNTRS-LoRA-X scores 81.61, and CNTDoRA-X scores 81.02. For CLIP-T, CNTRS-LoRA-X scores 27.86, and CNTDoRA-X scores 27.67. These integrations balance improvements across all metrics. Experiments on variable adapter structure (VAS) show performance gains when combined with CNTLoRA. CNTLoRA-S + VAS and CNTLoRA-Sh + VAS improves DINO, CLIP-I, and CLIP-T scores. VAS optimizes for variable rank allocation, enhancing overall performance. 
\begin{table}[h]
\caption{Cosine similarity and Spectral Norm between the initialized adapters and the final adapters for different methods. 
}
\vspace{- 1.5 em}
\label{tab:combined}
\begin{center}
\resizebox{0.45\textwidth}{!}
{
    \begin{tabular}{l|cccc|cccc}
    \toprule
& \multicolumn{4}{c|}{\textbf{Cosine Similarity} ($\uparrow$)}  & \multicolumn{4}{c}{\textbf{Spectral Norm} ($\downarrow$)}  \\ \midrule
\textbf{Method} & \textbf{Query} & \textbf{Key} & \textbf{Value} & \textbf{Out} & \textbf{Query} & \textbf{Key} & \textbf{Value} & \textbf{Out} \\ \midrule
LoRA & 0.515 & 0.356 & 0.366 & 0.522 & 4.490 & 3.953 & 1.907 & 1.408 \\
EVA & 0.583 & 0.492 & 0.434 & 0.783 & 0.614 & 0.627 & 1.022 & 0.791 \\ \midrule
\mycc CNTLoRA-X & \mycc \textbf{0.610} & \mycc \textbf{0.643} & \mycc \textbf{0.788} & \mycc \textbf{0.816} & \mycc \textbf{0.516} & \mycc \textbf{0.620} & \mycc \textbf{0.838} & \mycc \underline{0.646} \\
\mycc CNTLoRA-S & \mycc 0.519 & \mycc \underline{0.610} & \mycc \underline{0.623} & \mycc \underline{0.785} & \mycc \underline{0.517} & \mycc \underline{0.621} & \mycc 0.946 & \mycc 0.738 \\
\mycc CNTLoRA-Sh & \mycc \underline{0.586} & \mycc  0.535 & \mycc 0.613 & \mycc 0.760 & \mycc 0.574 & \mycc 0.726 & \mycc \underline{0.925} & \mycc \textbf{0.631} \\
    \bottomrule
    \end{tabular}}
\end{center}
\vskip -0.1in
\end{table}

\noindent \textbf{Difference between Initial and Final Weights} We evaluated the effectiveness of initialization methods by measuring the cosine similarity and spectral norm differences between initialized and fine-tuned adapters. Higher cosine similarity and lower spectral norm differences indicate better initialization, reducing weight adjustments during training. Table~\ref{tab:combined} shows that CNTLoRA variants consistently have higher similarity than EVA and LoRA, with CNTLoRA-X showing the strongest correlation. This suggests that our method starts adapters closer to their optimal configuration, leading to more faster convergence.

\begin{table}[h]
\caption{Final quantitative results i.e. DINO scores for (1000 iterations) on the Dreambooth dataset using SD 1.5 for different number of training samples per concept used in the initialization.}
\vspace{- 1.5 em}
\label{tab:gen15_ns}
\begin{center}
\resizebox{0.35\textwidth}{!}
{
    \begin{tabular}{l|cccc}
    \toprule
    \textbf{Method} & \textbf{n=1} & \textbf{n=2} & \textbf{n=4} & \textbf{All}
 \\ 
    \midrule
EVA & 61.23 & 61.26 & 61.84 & 62.24 \\
CORDA & 54.68 & 55.01 & 55.60 & 55.65 \\
LoRA-GA & 58.04 & 58.92 & 59.68 & 60.30 \\ \midrule
\mycc CNTLoRA-X & \mycc \textbf{64.02} & \mycc \textbf{64.28} & \mycc \textbf{64.50} & \mycc \textbf{64.63} \\
\mycc CNTLoRA-S & \mycc \underline{62.98} & \mycc \underline{63.16} & \mycc \underline{63.52}  & \mycc \underline{63.91} \\
\mycc CNTLoRA-Sh & \mycc 61.78 & \mycc {61.97} & \mycc {62.45} & \mycc 62.95 \\
\hline
    \bottomrule
    \end{tabular}}
\end{center}
\vskip -0.1in
\end{table} 

\noindent \textbf{Effect of Number of Samples} To assess the impact of the number of training samples on initialization, we fine-tuned Stable Diffusion v1.5 using EVA, CORDA, LoRA-GA, and CNTLoRA variants. This experiment reveals how each method initializes with varying amounts of data. We tested different sample sizes (n=1, 2, 4, All) to evaluate fine-tuning efficiency and model convergence. CNTLoRA-X consistently outperforms other methods, especially with fewer samples, highlighting the importance of high-quality initialization in low-data scenarios.

\begin{figure}[h]
    \centering
    \includegraphics[width=1.0\linewidth]{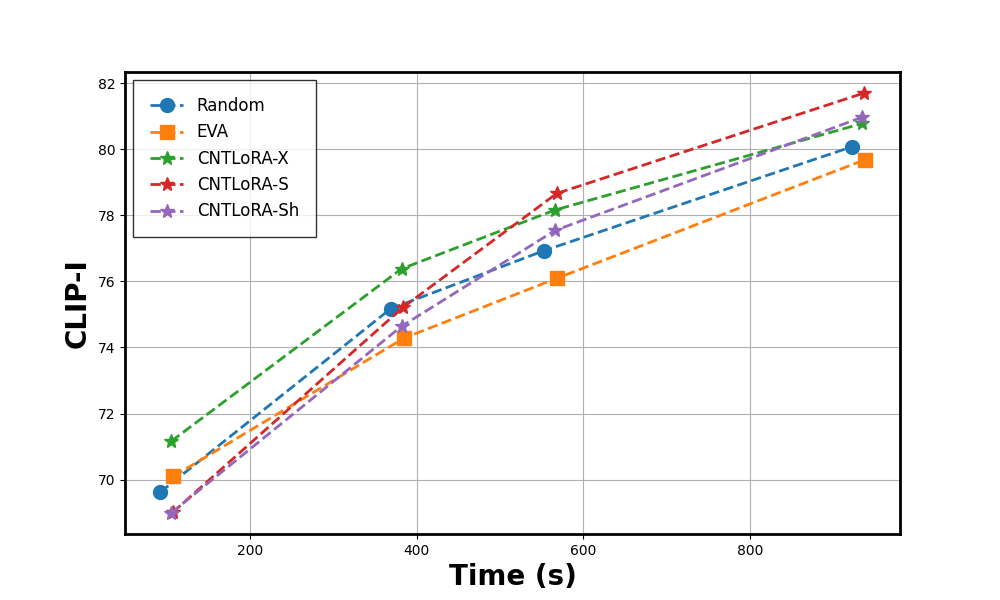}
    \vspace{- 1.5 em}
    \caption{Plot showing how the CLIP score evolves for different initialization methods as a function of wall clock time}
    \label{fig:img_conv_15_clipi}
    \vspace{- 1.25 em}
\end{figure}

\noindent \textbf{Convergence Study} We analyzed how different initialization strategies affect the fine-tuning dynamics by tracking the progression of CLIP-I score over 1000 training steps. Figure~\ref{fig:img_conv_15_clipi} shows CLIP-I score evolution for various methods, including EVA, OLoRA, PISSA, and our proposed CNTLoRA variants. The results show that CNTLoRA variants achieve a steeper increase in scores compared to baseline methods, highlighting their ability to accelerate convergence. In the later stages of training, CNTLoRA-X, CNTLoRA-S, and CNTLoRA-Sh consistently attain the highest CLIP-I scores, demonstrating their effectiveness.

\begin{table}[h]
\caption{Time (s) required for computing the initialization of different techniques while training over 1000 iterations (922.19 s) of a concept on the Dreambooth dataset using SD 1.5.}
\vspace{- 1.5 em}
\label{tab:init_times}
\begin{center}
\resizebox{0.47\textwidth}{!}
{

\begin{tabular}{l|ccccccc}
\toprule
\textbf{} & \textbf{PISSA} & \textbf{OLoRA} & \textbf{EVA} & \textbf{CORDA} & \mycc \textbf{CNTLoRA-X} & \mycc \textbf{CNTLoRA-S} & \mycc \textbf{CNTLoRA-Sh} \\ 
\midrule
\textbf{Init} ($\downarrow$) & 5.9 & 5.51 & 15.9 & 11.8 & \mycc 13.1 & \mycc 15.1 & \mycc 13.1 \\ 
\textbf{Training} & 922.19 & 922.19 & 922.19 & 922.19 & \mycc 922.19 & \mycc 922.19 & \mycc 922.19 \\ 
\textbf{\%} ($\downarrow$) & 0.64 & 0.59 & 1.71 & 1.27 & \mycc 1.42 & \mycc 1.63 & \mycc 1.41 \\ 
\bottomrule
\end{tabular}
}
\end{center}
\vspace{-2.0em}
\end{table} 

\noindent \textbf{Time for Initialization} In Table~\ref{tab:init_times}, we compare initialization times for various techniques over 1000 iterations on Dreambooth dataset using Stable Diffusion v1.5. OLoRA has the shortest initialization time at 5.51 seconds (0.59\% of total training time), followed by PISSA at 5.9 seconds (0.64\%). Our methods, CNTLoRA-X, CNTLoRA-S, and CNTLoRA-Sh, have competitive times of 13.1, 15.1, and 13.1 seconds, respectively. These results show balance between initialization efficiency and training performance, with OLoRA and PISSA being fastest, while CNTLoRA methods offer good trade-off.

\subsubsection{Image classification}

\begin{table}[h]
\caption{Fine-tuning DINOv2-g/14 on the VTAB-1K benchmark. We report average accuracy across five seeds. 
}
\label{tab:img_cls_main}
\vskip .1in
\centering
\setlength{\tabcolsep}{1.9pt}
\resizebox{0.5\textwidth}{!}{
\begin{tabular}{l|cccccc|cc|cccccc|c}
\multicolumn{1}{c|}{}&\multicolumn{6}{c|}{{Natural}}&\multicolumn{2}{c|}{{Specialized}}&\multicolumn{6}{c|}{{Structured}}&\\
&\multicolumn{1}{c}{\rotatebox[origin=c]{90}{Cifar100}}
&\multicolumn{1}{c}{\rotatebox[origin=c]{90}{Caltech101}}
&\multicolumn{1}{c}{\rotatebox[origin=c]{90}{DTD}}
&\multicolumn{1}{c}{\rotatebox[origin=c]{90}{Pets}}
&\multicolumn{1}{c}{\rotatebox[origin=c]{90}{SVHN}}
&\multicolumn{1}{c|}{\rotatebox[origin=c]{90}{Sun397}}
&\multicolumn{1}{c}{\rotatebox[origin=c]{90}{EuroSAT}}
&\multicolumn{1}{c|}{\rotatebox[origin=c]{90}{Retinopathy}}
&\multicolumn{1}{c}{\rotatebox[origin=c]{90}{CLEVR-Count}}
&\multicolumn{1}{c}{\rotatebox[origin=c]{90}{CLEVR-Dist}}
&\multicolumn{1}{c}{\rotatebox[origin=c]{90}{DMLab}}
&\multicolumn{1}{c}{\rotatebox[origin=c]{90}{KITTI-Dist}}
&\multicolumn{1}{c}{\rotatebox[origin=c]{90}{dSpr-Ori}}
&\multicolumn{1}{c|}{\rotatebox[origin=c]{90}{sNORB-Ele}}
&\multicolumn{1}{c}{\rotatebox[origin=c]{90}{Average}}
\\
\midrule
FFT        & 73.1 & 89.7 & 78.4  & 92.2 & 89.5 & 55.5  & 95.0 & 70.5 & 93.6 & 64.2 & 63.6 & 68.8 & 64.3 & 56.8 & 75.4\\ 
LoRA       & 85.9 & 92.2 & 82.2  & 94.5 & 64.1 & 63.6  & 92.6 & 76.6 & 97.7 & 65.3 & 62.1 & 83.6 & 63.0 & 52.3 & 76.8 \\ 
AdaLoRA    & 85.4 & 92.5 & 81.4  & 95.0 & 90.5 & 62.2  & 96.4 & 76.6 & 94.4 & 64.4 & 60.3 & 83.7 & 61.0  & 46.0 & 77.8 \\ 
PiSSA      & 85.5 & 93.6 & 82.3 & 94.6 & 92.8 & 62.3  & 96.6 & 76.3 & 95.0 & 66.3 & 63.2 & 84.9 & 60.1 & 48.6 & 78.7\\ 
OLoRA     & 85.5 & 93.0 & 82.1 & 95.1 & 78.3 & 62.1  & 96.3 & \underline{76.8} & 94.3 & 66.0 & 62.4 & 71.3 & 60.9 & 49.5 & 76.7 \\ 
EVA        & 85.6 & 93.9 & 82.2 & 95.9 & 93.2 & 63.6 & 96.6 & 76.1 & 96.1 & 65.1 & 61.1 & 83.3 & 61.6  & 55.0 & 79.2 \\ 
DoRA       & 85.9 & 92.7 & 82.1  & 95.2 & 34.4 & 61.4 & 96.8 & \underline{76.8} & 97.6 & 65.4 & 62.7 & 84.4 & 63.1 & 52.6 &  75.1 \\ 
EVA+DoRA  & \underline{86.2} & 92.1 & 81.9 & 94.9 & 93.8 & 62.4 & 96.6  & 76.7 & 97.2 & 65.5 & 54.1 & 83.7  & 62.3  & 54.5 & 78.7 \\
CoRDA      & 84.3 & 89.2 & 80.1     & 94.2 & 93.1 & 63.6    & 96.1     & 76.4 & 97.2 & 64.6 & 61.3 & 81.4     & 63.2     & 55.3 & 78.6 \\ \hline
\mycc CNTLoRA-X  & \mycc 86.1 & \mycc 94.0   & \mycc 83.1    & \mycc \underline{96.1} & \mycc 94.2 & \mycc 64.2    & \mycc \textbf{97.3}     & \mycc \textbf{77.0} & \mycc \underline{97.8} & \mycc 65.6 & \mycc 63.7 & \mycc 85.1     & \mycc \textbf{65.3}   & \mycc 57.0   &  \mycc 80.5\\ 
\mycc CNTLoRA-S  & \mycc \textbf{86.3} & \mycc \textbf{94.2} & \mycc \textbf{83.4}     & \mycc 96.0   & \mycc 94.1 & \mycc \textbf{64.4}     & \mycc \underline{97.1}     & \mycc \textbf{77.0} & \mycc \textbf{97.9} & \mycc \textbf{66.9} & \mycc \underline{64.7} & \mycc \textbf{85.3}    & \mycc \underline{65.2}     & \mycc \textbf{57.4} & \mycc  \textbf{80.7} \\
\mycc CNTLoRA-Sh & \mycc 86.1 & \mycc \underline{94.1} & \mycc \underline{83.2}     & \mycc \textbf{96.2} & \mycc \textbf{94.3} & \mycc \underline{64.3}    & \mycc 97.0     & \mycc \textbf{77.0} & \mycc 97.4 & \mycc \underline{66.4} & \mycc \textbf{64.8} & \mycc \underline{85.2}    & \mycc \underline{65.2}     & \mycc \underline{57.2} & \mycc \underline{80.6} \\
\bottomrule
\end{tabular}

}
\end{table}

\noindent \textbf{Quantitative Results} We fine-tuned the DINOv2-g/14 model on the VTAB-1K benchmark, which includes Natural, Specialized, and Structured datasets. Various initialization methods, including our CNTLoRA methods, were evaluated across five seeds for robustness. The results in Table \ref{tab:img_cls_main} show that CNTLoRA methods consistently achieve high performance. CNTLoRA-X, CNTLoRA-S, and CNTLoRA-Sh often outperform other methods like FFT, LoRA, and AdaLoRA. CNTLoRA-S achieved the highest average accuracy in several datasets, particularly excelling in the Natural category (e.g., Cifar100, Caltech101, DTD) and maintaining competitive performance in the Specialized and Structured categories (e.g., CLEVR-Count, CLEVR-Dist, DMLab).

\begin{figure}[t]
    \centering
    \includegraphics[width=1.0\linewidth]{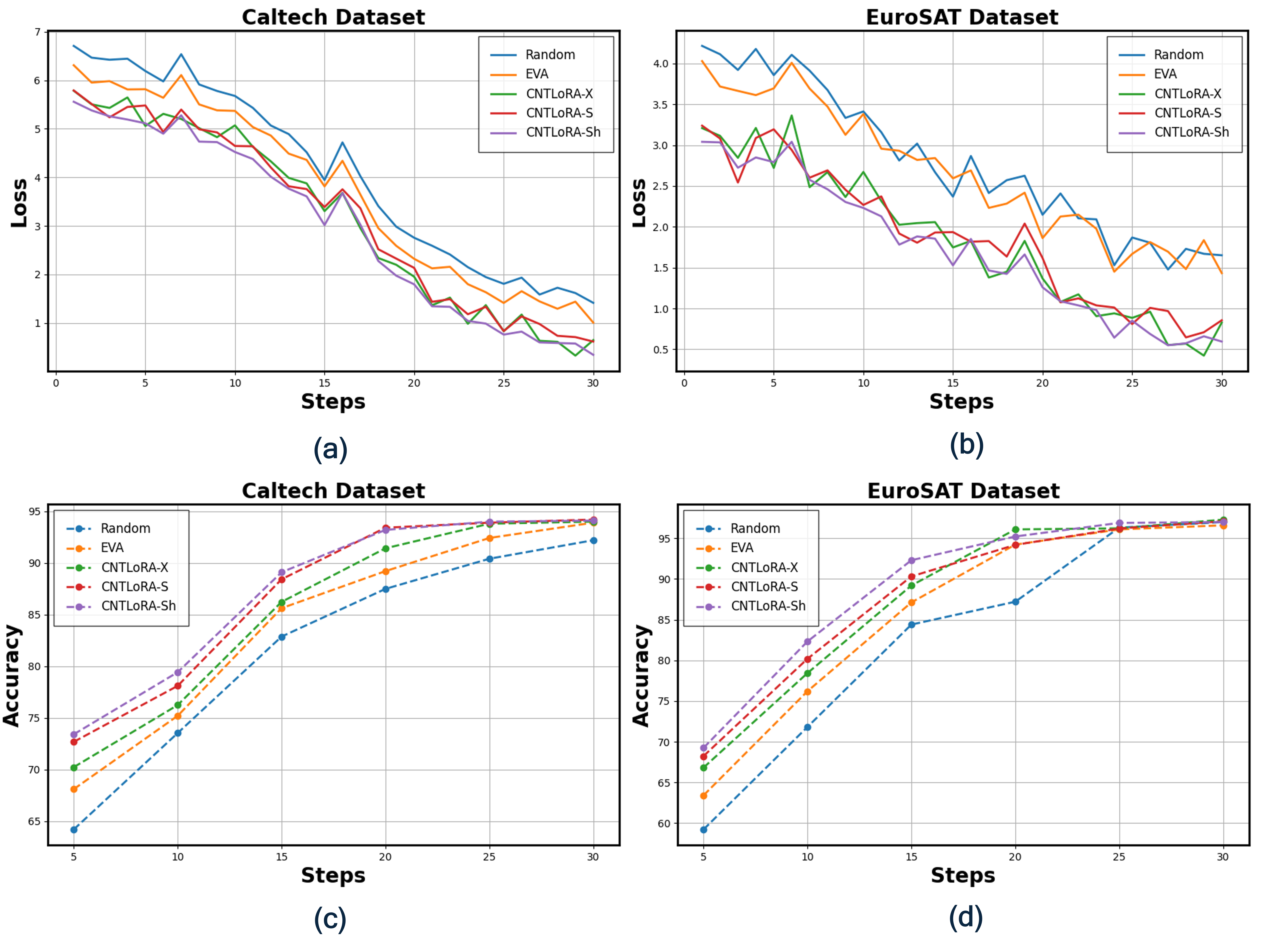}
    \vspace{- 1.5 em}
    \caption{Evolution of loss \& evaluation accuracy with epochs on the Caltech \& EuroSAT dataset}
    \label{fig:combined_score_cls}
    \vspace{- 1.25 em}
\end{figure}

\noindent \textbf{Convergence Study} Figures \ref{fig:combined_score_cls} illustrate the training dynamics of the DINOv2-g/14 model fine-tuned with various initialization methods on the Caltech101 and EuroSAT datasets. These plots provide insights into the convergence behavior and stability of the different methods. As shown in Figure \ref{fig:combined_score_cls}, the accuracy of the model on the Caltech101 dataset increases with the number of training steps. The CNTLoRA methods, particularly CNTLoRA-S, exhibit a faster convergence rate compared to other methods. CNTLoRA-S achieves higher accuracy earlier in the training process, indicating its effectiveness in LoRA fine-tuning. This rapid convergence suggests that CNTLoRA-S can effectively leverage the pre-trained features of the DINOv2-g/14 model, leading to improved performance. Figure \ref{fig:combined_score_cls} shows the accuracy progression on the EuroSAT dataset. Similar to the results on Caltech101, the CNTLoRA methods demonstrate superior performance. CNTLoRA-X and CNTLoRA-S consistently achieve higher accuracy throughout the training process. The plots indicate that these methods not only converge faster but also maintain higher accuracy levels compared to other methods.

\subsubsection{Image understanding}

\begin{table}[h]
\caption{Comparison on Image Understanding task for \textbf{Amazon Product Description (APD)} \& \textbf{myVLM (VLM)} datasets.
}
\label{tab:imu_main}
\begin{center}
\resizebox{0.45\textwidth}{!}
{
    \begin{tabular}{l|ccc}
    \toprule
    \textbf{Method} & \textbf{SentSim ($\uparrow$) (APD)} & \textbf{SentSim ($\uparrow$)(VLM)} & \textbf{Recall ($\uparrow$) (VLM)} \\ 
    \midrule
LoRA & 0.8013 & 0.6123 & 0.7134 \\
AdaLoRA & 0.8087 & 0.6164 & 0.7166 \\
PiSSA & 0.8132 & 0.6014 & 0.7143 \\
OLoRA & 0.8079 & 0.6115 & 0.7162 \\
EVA & 0.8186 & 0.6268 & 0.7250 \\
LoRA-GA & 0.8177 & 0.6194 & 0.7160 \\
CoRDA & 0.8121 & 0.6195 & 0.7180 \\
\midrule
\mycc CNTLoRA-X & \mycc 0.8117 & \mycc {0.6355} & \mycc 0.7320 \\
\mycc CNTLoRA-S & \mycc {0.8256} & \mycc {0.6385} & \mycc {0.7390} \\
\mycc CNTLoRA-Sh & \mycc {0.8273} & \mycc 0.6336 & \mycc {0.7341} \\
\midrule
\mycc CNTLoRA-X + VAS & \mycc 0.8192 & \mycc {0.6403} & \mycc 0.7419 \\
\mycc CNTLoRA-S + VAS & \mycc \underline{0.8301} & \mycc \textbf{0.6442} & \mycc \textbf{0.7495} \\
\mycc CNTLoRA-Sh + VAS & \mycc \textbf{0.8324} & \mycc \underline{0.6412} & \mycc \underline{0.7455} \\
\hline
    \bottomrule
    \end{tabular}
    }
\end{center}
\end{table} 

\noindent \textbf{Quantitative Results} The results summarized in Table \ref{tab:imu_main}, demonstrate that CNTLoRA methods consistently achieve high performance across different datasets. Table \ref{tab:imu_main} presents a comparison of various methods on image understanding task for the APD and VLM datasets. The evaluation metrics include SentSim (Sentence Similarity) for both datasets and Recall for VLM dataset. The best results are highlighted in bold, while the second-best results are underlined. The CNTLoRA methods, particularly CNTLoRA-S and CNTLoRA-Sh, demonstrate superior performance across metrics. For the APD dataset, CNTLoRA-Sh achieves the highest SentSim score of 0.8273, followed closely by CNTLoRA-S. This indicates that these methods are highly effective in understanding and relating visual content to textual descriptions. In the VLM dataset, CNTLoRA-S achieves the highest SentSim score of and the highest Recall score, showcasing its robustness and effectiveness in visual language model tasks. CNTLoRA-Sh also performs exceptionally well, with a SentSim score of 0.6336 and a Recall score of 0.7341, both of which are among top results. Compared to other methods, such as LoRA, AdaLoRA, PiSSA, OLoRA, EVA, LoRA-GA, and CoRDA, the CNTLoRA methods consistently achieve higher scores, indicating superior capability in image understanding tasks. Furthermore, as expected, when we apply VAS on top of all variations of CNTLoRA, the performance improves due to the optimal allocation of ranks.





\begin{figure}[h]
    \centering
    \includegraphics[width=1.0\linewidth]{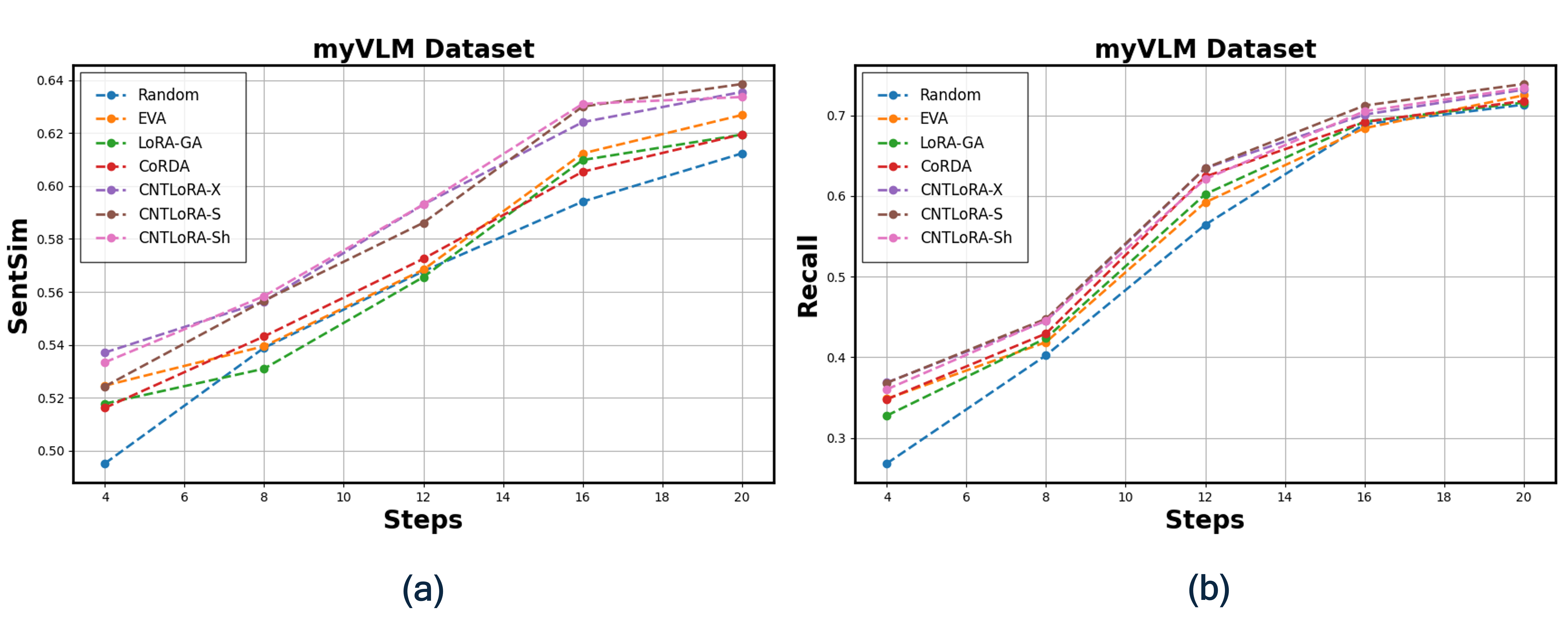}
    \vspace{- 1.5 em}
    \caption{Progression of evaluation metrics over epochs on myVLM dataset}
    \label{fig:combined_score}
\end{figure}

\noindent \textbf{Convergence Study} In Fig.~\ref{fig:combined_score}, we observe how Recall and SentSim values evolve with the increasing number of epochs on the myVLM dataset. Overall, our proposed CNTLoRA variations evolve faster and also produce much higher final convergence scores. On this dataset, data-driven LoRA initialization seems to be more effective, as shown by the poor performance of random initialization. Among our proposed variations CNTLoRA-S seems to be more effective suggesting that the covariance matching of activations is more important on this dataset.


\begin{figure}[t]
    \centering
    \includegraphics[width=1.0\linewidth]{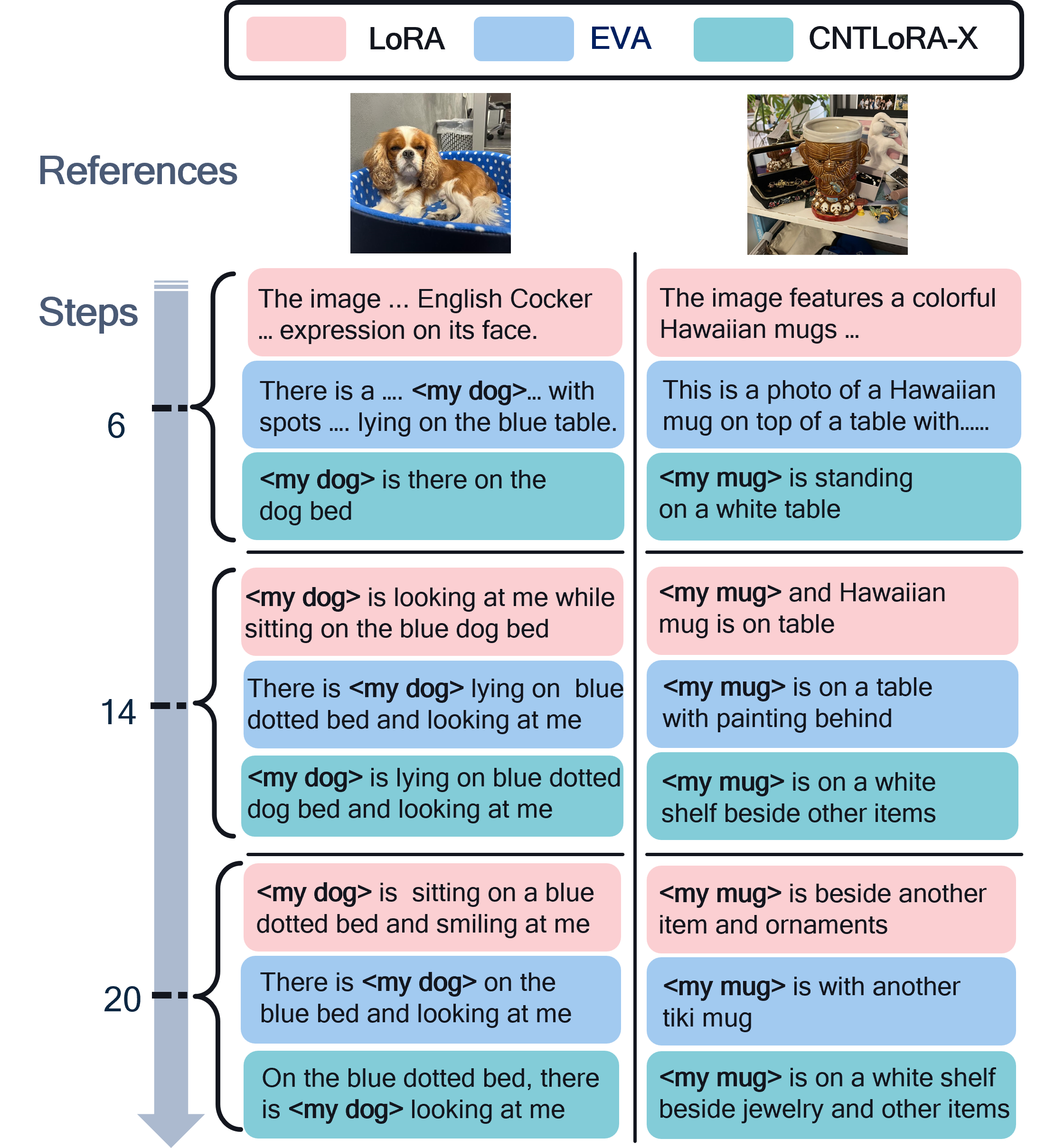} 
    \vspace{- 1.5 em}
    \caption{Figure showing how image captions evolve with increasing no. of steps for different input images. The samples are from myVLM dataset. Presence of $\mathbf{<}$\textbf{*}$\mathbf{>}$ in generated outputs suggests that the concept has been identified. Furthermore, generation of concise prompts suggest that model has been well fine-tuned.}
    \label{fig:imu_qual_conv}
\end{figure}

\noindent \textbf{Qualitative Results} Fig.~\ref{fig:imu_qual_conv} shows how captions on the evaluation set evolve for different training steps (i.e., 6, 14, 20) for different initialization methods: LoRA (random), EVA and CNTLoRA-X. For the $<$my dog$>$ case, LoRA cannot detect at 6 steps and hence produces a general caption. On the other hand, EVA and CNTLoRA-X can identify the presence of the concept. However, CNTLoRA-X produces a more appropriate response by identifying a dog bed instead of a table. For $<$my mug$>$, our proposed method can detect a personalized object at step 6, while other methods cannot. At step 20, our method produces more descriptive captions for mug compared to LoRA and EVA.

\section{Conclusion}

In conclusion, our data-driven weight initialization technique, CNTLoRA, significantly improves convergence speed and final performance of LoRA in fine-tuning tasks. By treating LoRA initialization as a domain shift problem and using activation vector constraints, we derived a closed-form estimate for LoRA weights. This enhances performance across tasks like image generation, classification, and understanding. Our analyses validate CNTLoRA’s robustness and efficacy, with CNTLoRA-X achieving better performance in few shot adaptation of the image generation task, while CNTLoRA-S producing better performance in the many-shot adaptation of image classification and understanding. Additionally, our proposed variable adapter structure improved recognition performance by allowing optimal rank allocation. Future work will explore more constraints to refine and expand our framework’s applicability.

\clearpage

\clearpage

{\small
\bibliographystyle{ieeenat_fullname}
\bibliography{biblo}
}

\clearpage

\appendix

\section{Additional Experimental Details}

\noindent \textbf{Image Generation} For the image generation task, we consider the Dreambooth~\cite{ruiz2023dreambooth} dataset. This dataset consists of 30 image sets from 15 different categories in which each category consists of 4 to 6 images per concept. The subjects primarily fall into two types: living or non-living. Based on this distinction, 25 prompts are used for evaluation across four different seeds. For the training, we used the default learning rate of 1e-4 (AdamW optimizer), with batch size of 1, with prior preservation loss weight of 0.1. The LoRA rank is set to 128 with $\alpha = 256$ and is applied to attention layers of the text encoder, as well as to both the self-attention and cross-attention layers of the UNet of Stable Diffusion V1.5. LoRA is attached to the key, query, value, and output projection matrices in the attention layers of both the UNet and the text encoder. Training is conducted over 1000 iterations per concept.

\noindent \textbf{Image Classification} For image classification, we fine-tune the {DINOv2-g/14} model~\cite{oquab2023dinov2} using {VTAB-1K~\cite{zhai2019large}}, which consists of 19 classification tasks spanning natural, specialized, and structured domains. In our experiments, we select 14 tasks from VTAB-1K to assess fine-tuning performance. To incorporate AdaLoRA, PiSSA, OLoRA, EVA, DoRA, and CoRDA, we adapt their implementations from the \texttt{peft} library. The classifier head is initialized with weights drawn from a normal distribution ($\sigma = 2e^{-5}$), while biases are set to zero. Throughout fine-tuning, we update the classification head, LoRA matrices, and biases. LoRA matrices are applied to most linear layers, especially query, key, and value components of attention layers, as well as the dense and fully connected layers. Input images are rescaled to $224 \times 224$ using bicubic interpolation and normalized according to ImageNet’s per-channel mean and variance. Fine-tuning is performed with bfloat16 precision, using AdamW (weight decay = 0.05) for 30 epochs. The learning rate follows a cosine decay schedule, with a linear warm-up phase spanning the first three epochs. For full fine-tuning, a layer-wise learning rate decay of 0.75 is applied.

\begin{figure}[h]
    \centering
    \includegraphics[width=1.0\linewidth]{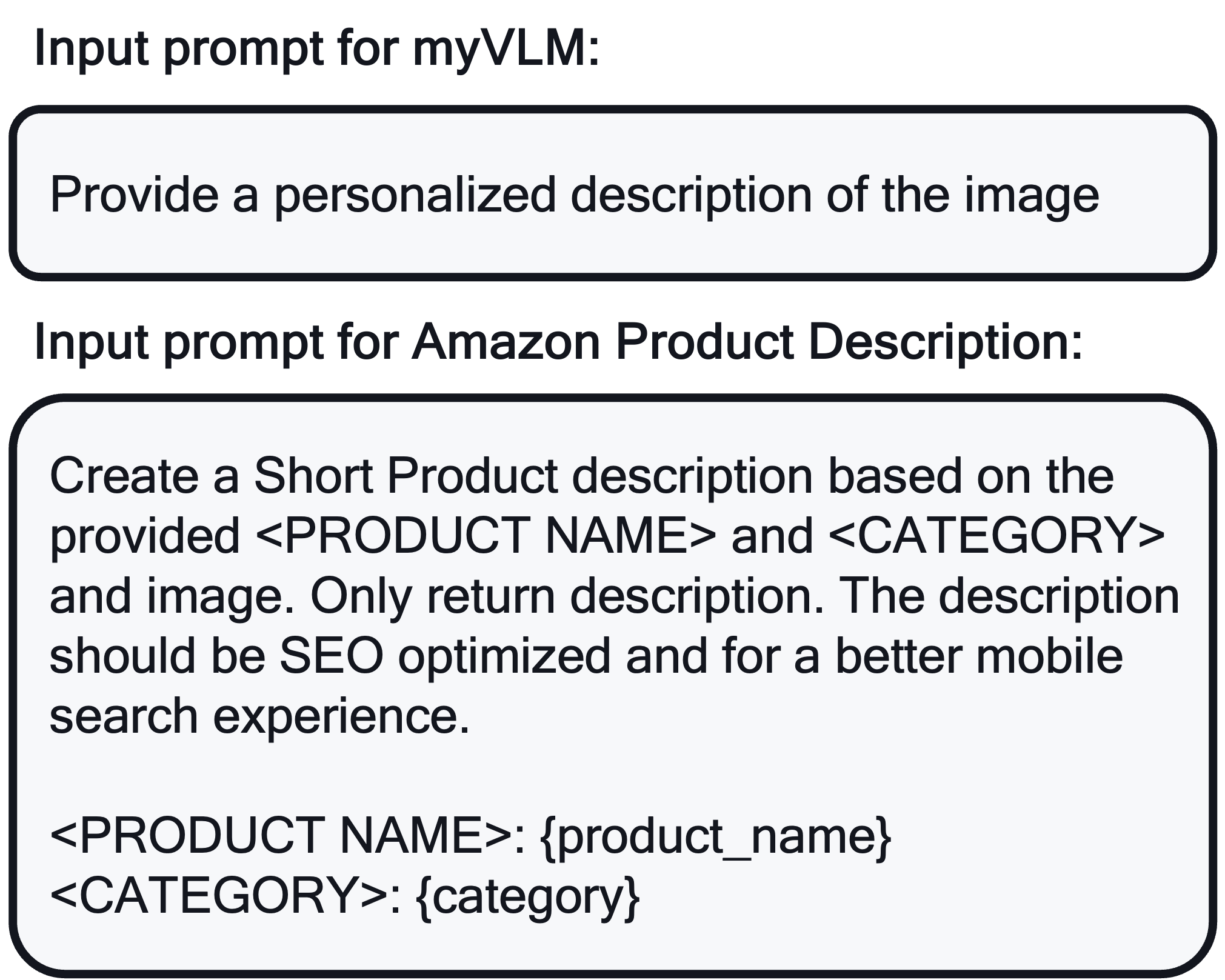} 
    \vspace{- 1.5 em}
    \caption{Figure showing the input prompts used in the myVLM and Amazon Product Description datasets.}
    \label{fig:imu_prompt}
    \vspace{- 1.25 em}
\end{figure}

\noindent \textbf{Image Understanding} For image understanding, we consider {Amazon Product Description dataset (APD)}~\cite{amznproddesc} for product marketing. The APD dataset is a large-scale resource designed for product marketing and e-commerce applications, encompassing both structured metadata and unstructured textual descriptions across a wide range of product categories. For fine-tuning, we use the {GLM-Edge} model~\cite{glm2024chatglm}, 
incorporating LoRA-based adaptation with a rank of 32 and $\alpha = 64$, optimizing it to generate domain-specific product descriptions. The attachment points are done at query, key and value locations of all attention layers in the vision encoder as well as the language model.
Additionally, we utilize the {myVLM} dataset~\cite{alaluf2024myvlm}, which is tailored for personalized vision-language modeling, focusing on concept-based captioning where descriptions are tailored to user-defined preferences and contexts. The dataset comprises manually annotated image-text pairs, ensuring high-quality supervision for customized caption generation and multimodal retrieval tasks. We fine-tune {LLAVA-v1.5-7B}~\cite{liu2023visual}, a vision-language model, applying adapters with a rank of 128 and $\alpha=256$ to the self-attention layers in the visual encoder and both the self-attention and feed-forward network (FFN) layers in the language model. For the attention blocks, LoRA is attached at the query, key and value locations. In both the models, training is conducted using AdamW optimizer with a learning rate of 2e-4 and batch size of 8. For the both tasks, fine-tuning is performed over 20 epochs using all available samples. Fig.~\ref{fig:imu_prompt} shows the input prompts used in the datasets.

\noindent \textbf{Language Understanding} For language understanding, we consider the popular {GLUE} Benchmark~\cite{wang2018glue}. This benchmark consists of eight downstream tasks, such as natural language inference, or sentiment analyses. For the base model, we use the large version of {Roberta}~\cite{liu2019roberta}. The hyperparameters used are the same as EVA~\cite{paischer2024one}. For the evaluation metrics, we report accuracy for all tasks except for CoLA and STS-B, where we report Matthew's correlation and Pearson's correlation, respectively.

\section{Gradient Analysis}
\label{sec:grad_info}

Our proposed initialization method requires initialization of $\BA^{i}$, the down matrices and  $\BB^{i}$, the up matrices separately. The gradients of $\BA^{i}$ and $\BB^{i}$ with respect to the task loss $L$ is given as $\frac{\partial L}{\partial \BA^{i}} = \BB^{iT} \left( \frac{\partial L}{\partial \BY} \right) \BX^{iT}_{tar}$, where $\BY$ is the output of the LoRA layer. Similarly, the gradient of the task loss $L$ with respect to $\BB^{i}$ is given as $\frac{\partial L}{\partial \BB^{i}} =\left( \frac{\partial L}{\partial \BY} \right) \BX^{iT}_{tar}  \BA^{iT}$. From the expression, it is clear that the initial gradients would depend on the initial values of $\BA^{i}$ and $\BB^{i}$. For random initialization, $\frac{\partial L}{\partial \BA^{i}}$ would be zero as $\BB^{i}$ is initialized to $\mathbf{0}$. This can slow down the convergence. For data-driven initialization like EVA, $\BA^{i}$ is initialized from data and it is likely to assist $\frac{\partial L}{\partial \BB^{i}}$ in the convergence. However, still $\BB^{i}$ is initialized to 0 and hence the initial gradients of $\frac{\partial L}{\partial \BA^{i}}$ would be zero and hence affecting convergence. For our proposed method, the initial values of $\BB^{i}$ and $\BA^{i}$ are non-zero and both depend on the principal components of the input activations $\BX^{i}_{tar}$. Consequently, both of the gradients have a dependency on $\BX^{i}_{tar}$ beyond first order and hence we expect faster convergence compared to EVA.

\section{Competitive Low Rank Adaptors as Constraints}

The first example of \textbf{Native LoRA} is as follows:
\begin{align}\label{eq:type1}
&\BF_1(\BX^{i}_{src}, \BX^{i}_{tar}, \BW^{i}_{src}, \BW^{i}_{tar}) = \mathbf{0}  \\
&\BW^{i}_{tar}  - \BW^{i}_{src} = \mathbf{0} \implies \Delta \BW^{i} = \mathbf{0} \\
&\implies \BB^{i} = \mathbf{0}, \BA^{i} \sim \mathcal{N}(\mathbf{0}, \BI) 
\end{align}
\textbf{PISSA} can be expressed as: 
\begin{align}\label{eq:type2}
&\BF_1(\BX^{i}_{src}, \BX^{i}_{tar}, \BW^{i}_{src}, \BW^{i}_{tar}) = \mathbf{0}\\
&\BW^{i}_{tar}  - 2\BW^{i}_{src} = \mathbf{0} 
\implies \Delta \BW^{i} = \BW^{i}_{src} \\
&\Delta\BW^{i} = \BU^{i} \BS^{i} \BV^{i} \implies \BB^{i} = \BU^{i}[::r] \BS^{i}[:r]^{0.5} , \\
&\BA^{i} = \BS^{i}[:r]^{0.5} \BV^{i}[:r:]  
\end{align}
\textbf{OLORA} can be expressed as: 
\begin{align}\label{eq:type3}
&\BF_1(\BX^{i}_{src}, \BX^{i}_{tar}, \BW^{i}_{src}, \BW^{i}_{tar}) = \mathbf{0} \\
&\BW^{i}_{tar}  - 2\BW^{i}_{src} = \mathbf{0} 
\implies \Delta \BW^{i} = \BW^{i}_{src} \\
&\Delta\BW^{i} = \BQ^{i}[::r] \BR^{i}[:r:] \implies \BB^{i} = \BQ^{i}[::r]  , \\
&\BA^{i} = \BR^{i}[:r:] 
\end{align}
\textbf{CORDA} can be expressed as: 
\begin{align}\label{eq:type4}
&\BF_1(\BX^{i}_{src}, \BX^{i}_{tar}, \BW^{i}_{src}, \BW^{i}_{tar}) = \mathbf{0}\\
&(\BW^{i}_{tar}  - \BW^{i}_{src})\BX^{i}_{tar}\BX^{iT}_{tar} - \text{SVD}((\BW^{i}_{tar})\BX^{i}_{tar}\BX^{iT}_{tar}) = \mathbf{0} \\
&\implies \Delta \BW^{i}\BC^{i}_{tar} = \text{SVD}(\BW^{i}_{tar}\BC^{i}_{tar}) \\
&\implies  \Delta \BW^{i} = \text{SVD}(\BW^{i}_{tar}\BC^{i}_{tar})\BC{^{i}_{tar}}^{-1}  \\
&\implies \BA^{i} = (\BS^{i})^{0.5}[:r](\BV^{i}\BC^{i}_{tar})^{-1}[:r:] \\
&\implies \BB^{i} = \BU^{i}[::r] (\BS^{i})^{0.5}[:r]
\end{align}
\textbf{EVA} can be expressed as: 
\begin{align}\label{eq:type5}
&\BF_1(\BX^{i}_{src}, \BX^{i}_{tar}, \BW^{i}_{src}, \BW^{i}_{tar}) = \mathbf{0}\\
&\BA^{i} = \BV^{i}[:r:] \quad \text{where} \\
&\BU^{i}[::r],\BS^{i}[:r],\BV^{i}[:r:] = \text{SVD}(\BX^{i}_{tar})  
\end{align}

\section{Additional Experiments}
\subsection{Image generation}
\noindent \textbf{Effect of different ranks}
We also study how different ranks affect quantitative performance of the model as shown in Table~\ref{tab:rank}. We have originally used the default rank of 128, on which our proposed method and their variations CNTLoRA-X, CNTLoRA-S, CNTLoRA-Sh produced better DINO scores than competitive methods. The pattern holds true even for lower ranks of 64 and 32. In fact, CNTLoRA-X at rank 64 produces better DINO scores than what LoRA and EVA produces at rank of 128.
\begin{table}[h]
\caption{Final quantitative results (DINO) (1000 iterations) of different methods on the Dreambooth dataset using SD 1.5 for different ranks. 
Higher ($\uparrow$)  is better.
}
\vspace{- 1.5 em}
\label{tab:rank}
\begin{center}
\resizebox{0.3\textwidth}{!}
{
    \begin{tabular}{l|ccc}
    \toprule
    \textbf{Method} & \textbf{128}  & \textbf{64} & \textbf{32} 
 \\ 
    \midrule
LoRA & 62.74 & 59.07 & 56.62 \\
EVA & 62.24 & 60.93 & 57.41 \\
\midrule
\mycc CNTLoRA-X & \mycc {64.63} & \mycc 62.98 & \mycc 59.09 \\
\mycc CNTLoRA-S & \mycc {63.91} & \mycc {62.25} & \mycc 58.23 \\
\mycc CNTLoRA-Sh & \mycc 62.95 & \mycc {61.02} & \mycc {57.69} \\  
\hline
    \bottomrule
    \end{tabular}}
\end{center}
\vspace{- 1.5 em}
\end{table} 

\begin{figure*}[h]
    \centering
    \includegraphics[width=1.0\linewidth]{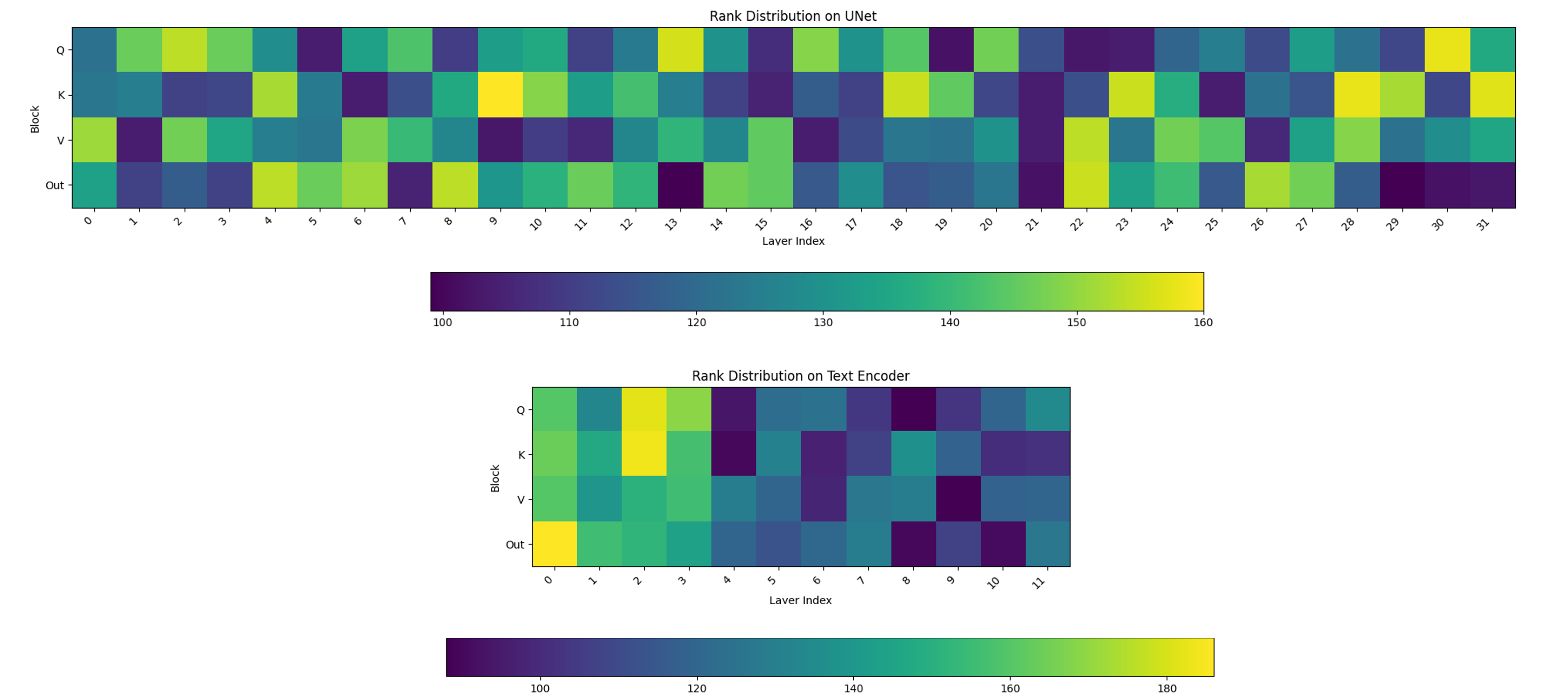}
    \caption{Plot showing how the rank distribution takes place for the dog class in the Dreambooth dataset.}
    \label{fig:vas-map}
\end{figure*}

\noindent \textbf{Effect of different learning rates}
We also study how different learning rates affect quantitative performance of the model as shown in Table~\ref{tab:lr}. We have originally used the default learning rate of 1e-4, on which our proposed method and their variations CNTLoRA-X, CNTLoRA-S, CNTLoRA-Sh produce better DINO scores than competitive methods. The pattern holds true even for different learning rates of 1e-3 and 1e-5. For both these learning rates we see a general drop in performance compared to that at 1e-3. The reason for drop at 1e-5 is because of slower learning rate and not enough iterations to converge. The reason for drop in learning rate at 1e-3 is due to instability in the fine-tuning procedure.
\begin{table}[h]
\caption{Final quantitative results (DINO) (1000 iterations) of different methods on the Dreambooth dataset using SD 1.5 for different learning rates. 
}
\vspace{- 1.5 em}
\label{tab:lr}
\begin{center}
\resizebox{0.3\textwidth}{!}
{
    \begin{tabular}{l|ccc}
    \toprule
    \textbf{Method} & \textbf{1e-4}  & \textbf{1e-3} & \textbf{1e-5} 
 \\ 
    \midrule
LoRA & 62.74 & 60.51 & 58.42 \\
EVA & 62.24 & 60.26 & 59.01 \\
\midrule
\mycc CNTLoRA-X & \mycc {64.63} & \mycc 61.98 & \mycc 60.05 \\
\mycc CNTLoRA-S & \mycc {63.91} & \mycc {61.92} & \mycc 60.22 \\
\mycc CNTLoRA-Sh & \mycc 62.95 & \mycc {60.81} & \mycc 59.10 \\  
\hline
    \bottomrule
    \end{tabular}}
\end{center}
\vspace{- 1.5 em}
\end{table} 

\noindent \textbf{Rank distribution in VAS}
In our proposed Variable Adapter Structure (VAS) framework, we have variable ranks across different attachment points. In Fig.~\ref{fig:vas-map}, we visualize this variable rank allocation across attachment points for both the UNet and Text Encoder when finetuned on the dog class of the Dreambooth dataset. For the UNet case, we see that the rank is distributed more or less randomly across all attachment points. For the "Out" attachment point in the attention block, we see lower rank allocation in deeper layers suggesting lesser importance of LoRA modules in that region.  For the text encoder case, we see higher rank allocation in the initial layers while lower rank allocation in the deeper layers of the text encoder. This suggests lesser importance of LoRA modules in the deeper layer.

\begin{table}[h]
\caption{Final quantitative results (1000 iterations) of different variations of our method on the Dreambooth dataset using SD 1.5.}
\label{tab:diff_var}
\begin{center}
\resizebox{0.45\textwidth}{!}
{
    \begin{tabular}{l|ccc}
    \toprule
    \textbf{Method} & \textbf{DINO} ($\uparrow$) & \textbf{CLIP-I} ($\uparrow$) & \textbf{CLIP-T} ($\uparrow$)
 \\ 
    \midrule
\mycc CNTLoRA-X-QR & \mycc \underline{65.29} & \mycc 78.63 & \mycc 27.36 \\
\mycc CNTLoRA-X-0.75 & \mycc \textbf{65.78} & \mycc \textbf{81.34} & \mycc 24.91 \\
\mycc CNTLoRA-X-0.25 & \mycc 63.26 & \mycc 79.71 & \mycc 27.01 \\
\mycc CNTLoRA-X-0.1 & \mycc 62.41 & \mycc 78.23 & \mycc \textbf{28.34} \\
\mycc CNTLoRA-Sh-Norm. & \mycc 61.2 & \mycc 79.34 & \mycc 28.10 \\
\mycc CNTLoRA-Sh-10 & \mycc 62.13 & \mycc 79.92 & \mycc \underline{28.13} \\
\mycc CNTLoRA-Sh-0.1 & \mycc 62.42 & \mycc \underline{79.96} & \mycc 27.95 \\

\hline
    \bottomrule
    \end{tabular}}
\end{center}
\vskip -0.1in
\end{table}

\begin{figure*}[h]
    \centering
    \includegraphics[width=1.0\linewidth]{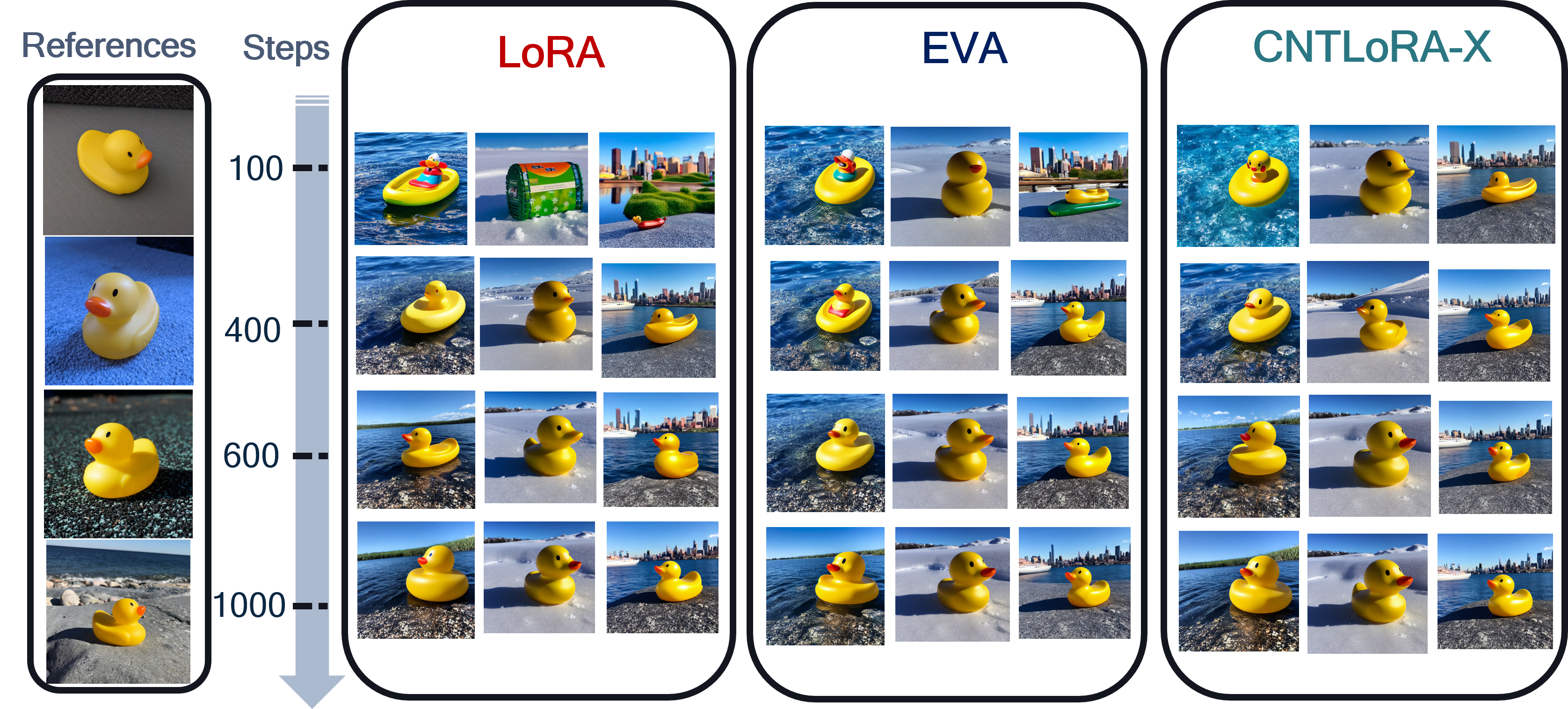}
    \vspace{- 1.5 em}
    \caption{Plot showing how the generation evolves for different initialization methods on the Dreambooth dataset for the duck class. The prompts from left to right are (a) A S* toy floating on top of water. (b) A S* toy in the snow. (c) A S* toy with a city in the background.}
    \label{fig:duck}
\end{figure*}

\begin{figure*}[h]
    \centering
    \includegraphics[width=1.0\linewidth]{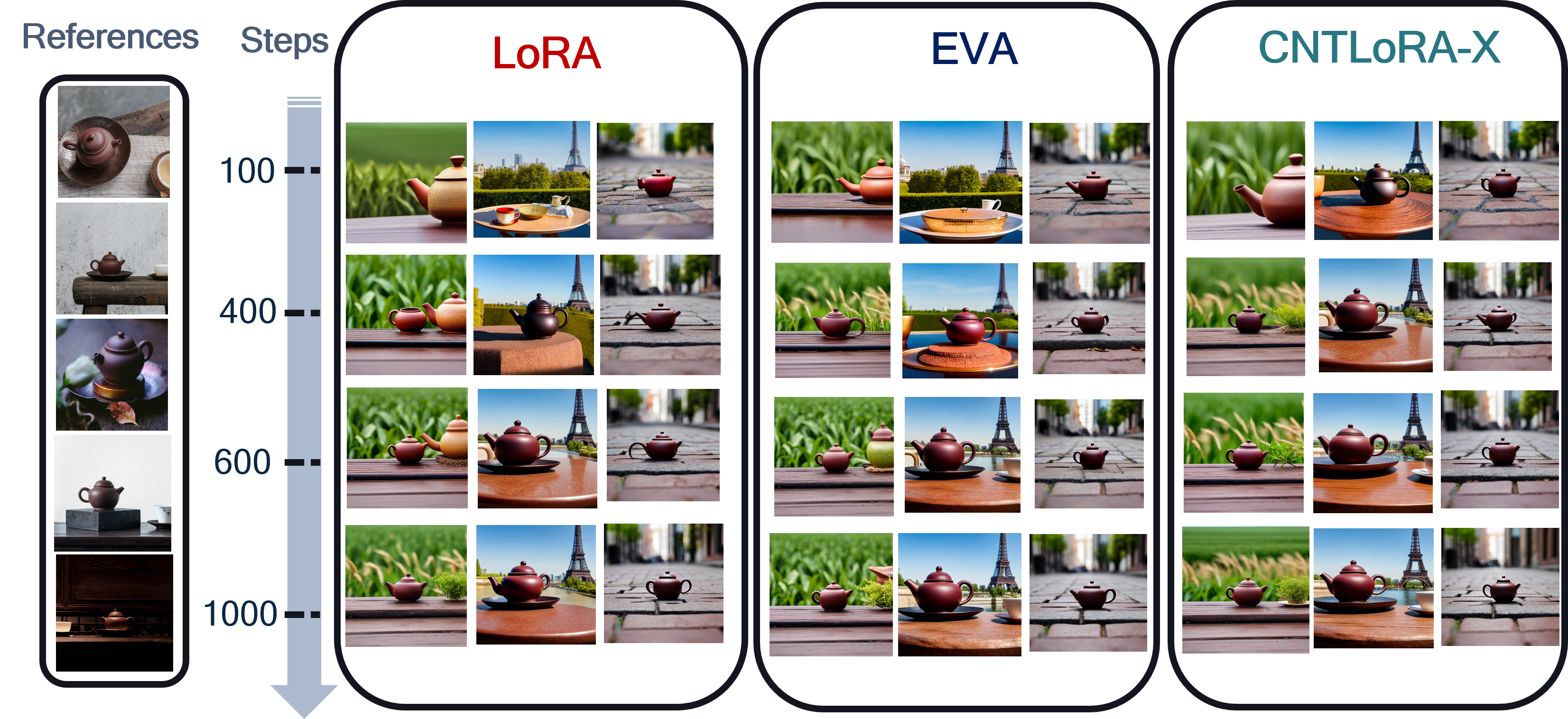}
    \vspace{- 1.5 em}
    \caption{Plot showing how the generation evolves for different initialization methods on the Dreambooth dataset for the teapot class. The prompts from left to right are (a) A S* teapot with a wheat field in the background. (b) A S* teapot with the Eiffel tower in the background. (c) A S* teapot on a cobblestone street.}
    \label{fig:teapot}
\end{figure*}

\begin{figure*}[h]
    \centering
    \includegraphics[width=1.0\linewidth]{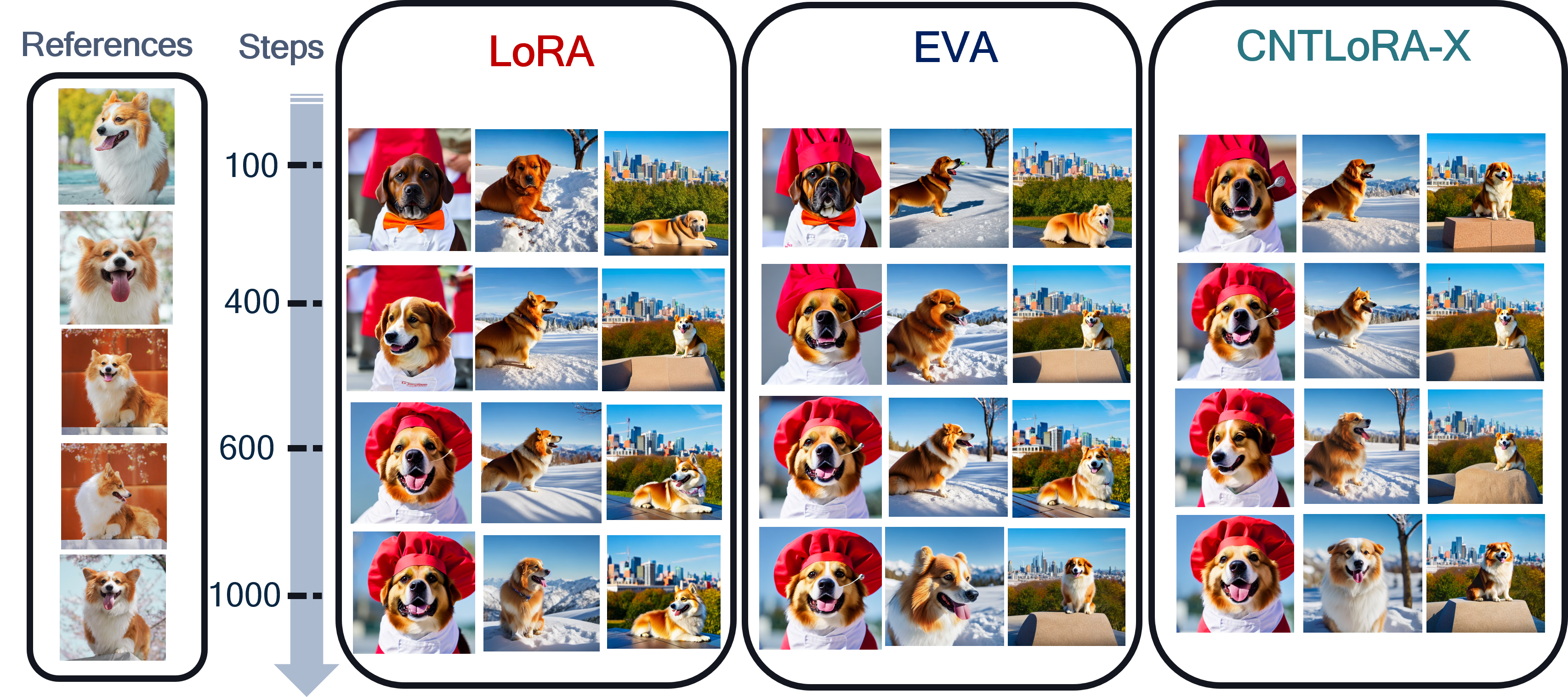}
    \vspace{- 1.5 em}
    \caption{Plot showing how the generation evolves for different initialization methods on the Dreambooth dataset for the dog class. The prompts from left to right are (a) A S* dog in a chef outfit. (b) A S* dog in the snow. (c) A S* dog with a city in the background.}
    \label{fig:dog}
\end{figure*}

\noindent \textbf{Additional Variations} In Table~\ref{tab:diff_var}, we consider different variations of CNTLoRA. CNTLoRA-X-QR uses QR decomposition instead of SVD for allocating the up and down matrices. CNTLoRA-X-$p$ uses fractional allocation of $p$ when splitting the singular matrix $\BS$ and allocating it to up and down matrices. By default,  we use $p=0.5$. Furthermore, we consider different variations of CNTLoRA-Sh where the constant $\BC$ is varied from the default value of identity: (a) \emph{Norm} where $\BC$ is normally distributed. (b) \emph{10} where $\BC$ is 10 times identity (c) \emph{0.1} where $\BC$ is 0.1 times identity. Overall, we see that the DINO score and CLIP-T score of CNTLoRA-X-QR improves over CNTLoRA-X. This might be due to the fact that for QR decomposition, there is no fractional allocation of singular matrix between up and down matrices and hence there is balance between image and text fidelity quite well. In fact, the fractional value $p$ can allow us to balance image and text fidelity well. As seen for higher fraction value of 0.75, DINO score increases while CLIP-T score decreases. If we decrease the fraction value to 0.25, DINO score decrease to 63.26 while the CLIP-T score increases to 27.01. When the fraction value is decreased further to 0.1, DINO score decreases to 62.41 while CLIP-T increases to 28.04. Hence, our proposed fractional allocation can maintain a balance between image and text fidelity. As for different variants of CNTLoRA-Sh, the newer hyperparameters of $\BC$ seems to produce poorer image fidelty performance. However, they lead to better prompt fidelity.

\noindent \textbf{Qualitative Results} We also show qualitative results on additional prompts for the duck, teapot and dog classes in Figs.~\ref{fig:duck},~\ref{fig:teapot} and ~\ref{fig:dog} respectively. In Fig.~\ref{fig:duck}, we observe that at 100 iterations of training, LoRA (i.e. random initialization) produces poor performance in terms of image fidelity. However, for both EVA and CNTLoRA-X, we observe better image fidelity due to data-driven initialization. Infact, for the prompt ``A S* toy floating on top of water", CNTLoRA-X produces better image fidelity where the face of the duck atleast appears compared to that of EVA.

In Fig.~\ref{fig:teapot}, we observe that at 100 iterations of training, LoRA (i.e. random initialization) and EVA produces poor performance in terms of image fidelity for the prompt ``A S* teapot with the Eiffel tower in the background.". However, for CNTLoRA-X, we observe better image fidelity. The pattern is repeated even for the prompt ``A S* teapot with a wheat field in the background."

In Fig.~\ref{fig:dog}, we observe that at 100 iterations of training, LoRA (i.e. random initialization) and EVA produces poor image fidelity performance for the prompts ``A S* dog in a chef outfit." and ``A S* dog with a city in the background.". However, for CNTLoRA-X, we observe better image fidelity. Even at 400 iterations of training, CNTLoRA-X produces better image fidelity for the prompt ``A S* dog in the snow".

\begin{figure}[h]
    \centering
    \includegraphics[width=1.0\linewidth]{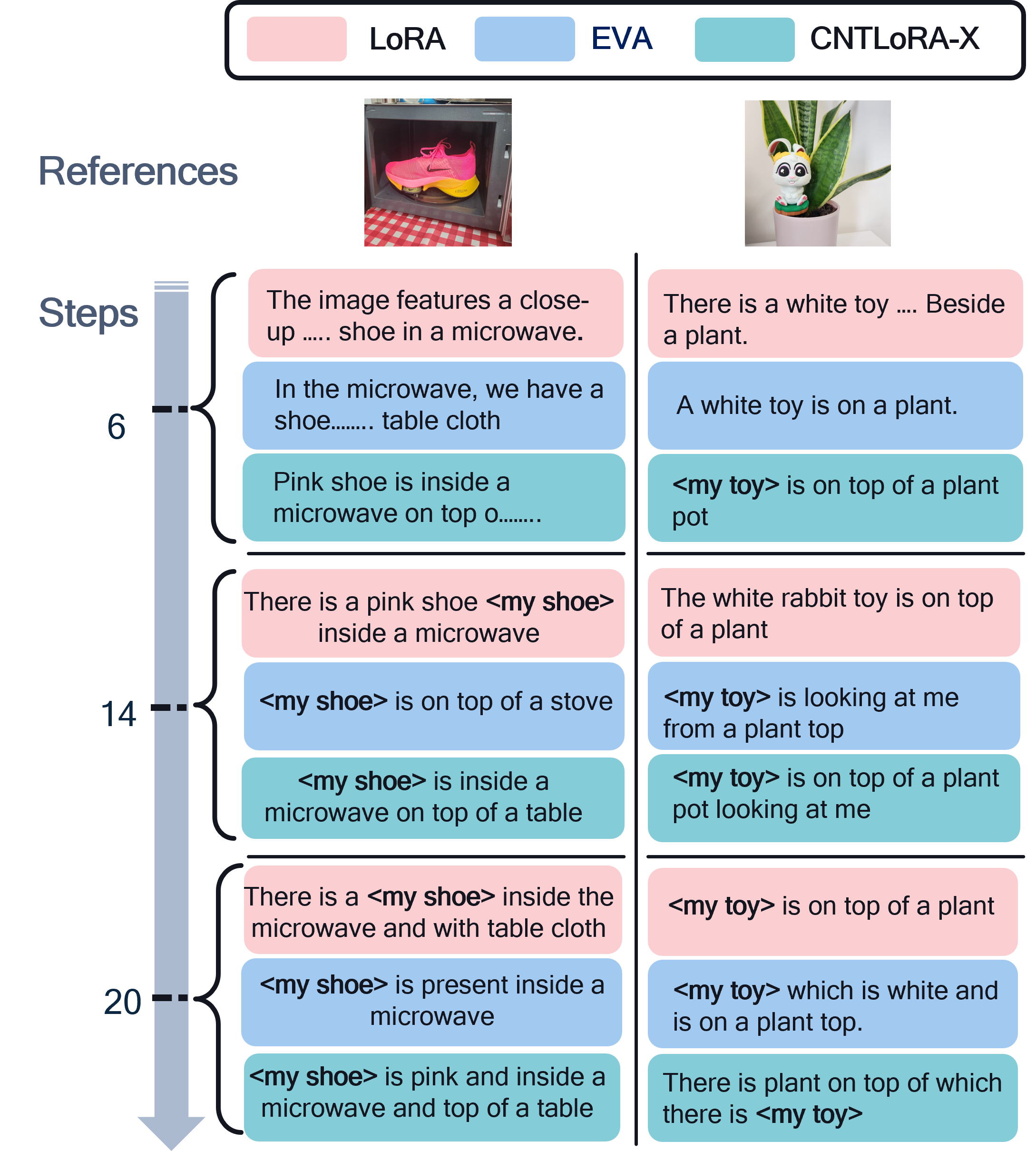} 
    \vspace{- 1.5 em}
    \caption{Figure showing how image captions evolve with increasing number of steps for different conditioning input images. The samples are from myVLM dataset. Presence of $\mathbf{<}$\textbf{*}$\mathbf{>}$ in the generated outputs suggest that the concept has been identified. Furthermore, generation of concise prompts suggest that the model has been well fine-tuned.}
    \label{fig:imu_qual_conv_supp}
    \vspace{- 1.25 em}
\end{figure}

\begin{figure*}[h]
    \centering
    \includegraphics[width=1.0\linewidth]{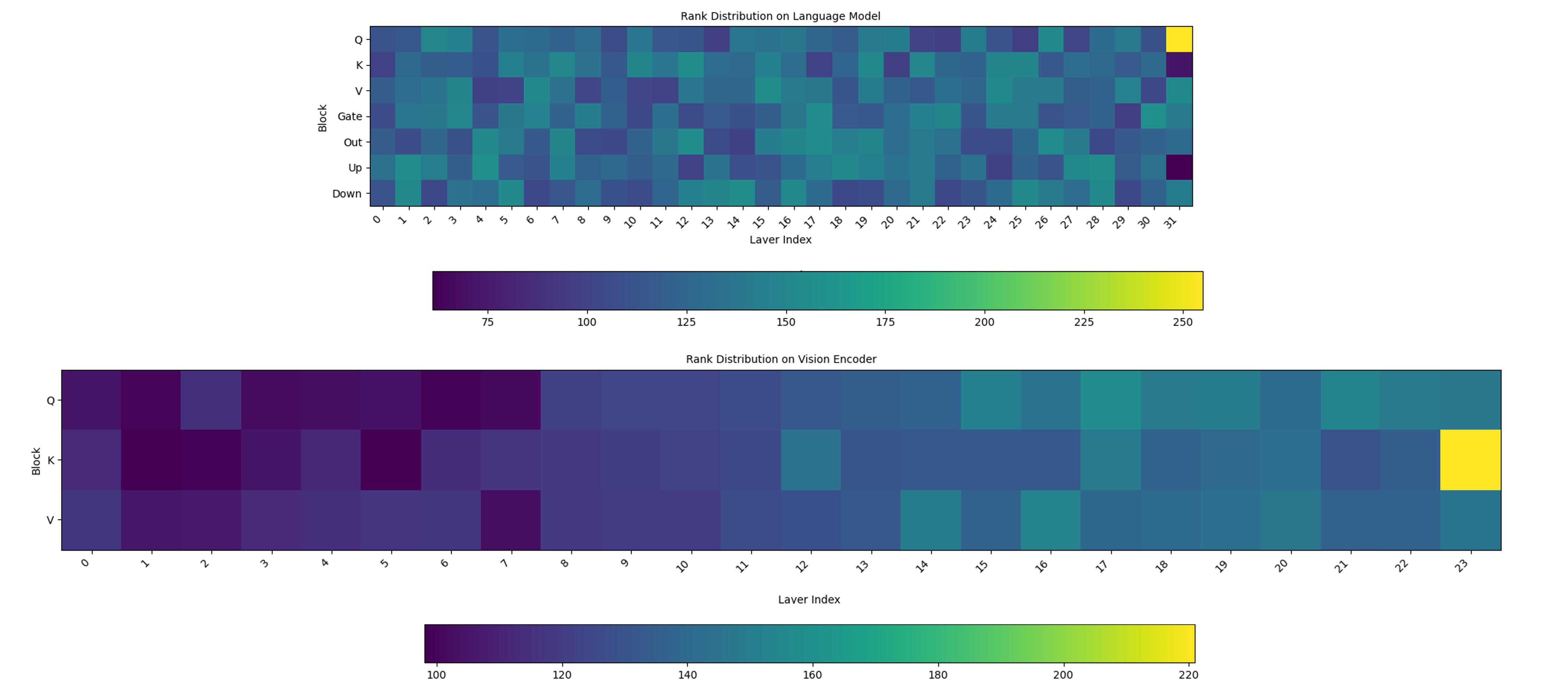}
    \caption{Plot showing how the rank distribution takes place when fine-tuned on the myVLM dataset. }
    \label{fig:vas-map-vlm}
\end{figure*}

\subsection{Image Understanding}

\noindent \textbf{Qualitative results} In Fig.~\ref{fig:imu_qual_conv_supp}, we observe how the captions evolve for different training steps i.e. 6, 14, 20 for the $<$my shoe$>$ and $<$my toy$>$ object. Our proposed framework CNTLoRA-X can recognize the personalized toy even at 6 steps. Even at 20 steps, it produces a more descriptive caption compared to LoRA and EVA. For the personalized shoe, our proposed method can produce more descriptive and accurate captions at Step 14 and Step 20. At step 6, even though the personalized object is not identified, our method CNTLoRA-X produces more accurate descriptions.

\noindent \textbf{Rank Distribution in VAS}
In our proposed Variable Adapter Structure (VAS) framework, we have variable ranks across different attachment points. In Fig.~\ref{fig:vas-map-vlm}, we visualize this variable rank allocation across attachment points for both the language model and vision encoder when fine-tuned on the myVLM dataset. For the language model case, we see that the rank is distributed more or less randomly across all attachment points. For the vision encoder case, we see higher rank allocation in the later layers while lower rank allocation in the shallower layers of the text encoder. This suggests that the fine-tuning image dataset has not very different distribution from pre-training dataset. Rather, the language style is changed and adapters need to be learned mainly for the language model.

\begin{table*}[h]
\caption{Comparison of all methods for \roberta{}~\cite{liu2019roberta} on GLUE tasks. We report Matthew's correlation for CoLA, Pearson correlation for STS-B, matched accuracy for MNLI, and accuracy for remaining tasks.}
\label{tab:glue_main}
\setlength{\tabcolsep}{5pt}
\begin{center}
\resizebox{0.7\textwidth}{!}{
    \begin{tabular}{l|ccccccccc}
    \toprule
   \textbf{ Method} & \textbf{MNLI} & \textbf{QNLI} &\textbf{ QQP }& \textbf{SST2} & \textbf{CoLA} &\textbf{ MRPC }& \textbf{RTE} & \textbf{STS-B} & \textbf{Avg}\\ 
    \midrule
    FFT                & $90.2$       & $94.7$      & $92.2$       & $\underline{96.4}$       & $68.0$       & $90.9$       & $86.6$       & $92.4$       & $88.93$ \\
    LoRA               & $90.7$       & $94.8$      & $92.0$       & $96.2$                  & $69.1$       & $91.1$       & $88.1$       & $92.3$       & $89.29$ \\
    AdaLoRA            & $90.5$       & $94.8$      & $90.6$       & $96.1$                  & $68.2$       & $90.7$       & $84.4$       & $91.8$       & $88.39$ \\
    PiSSA              & $90.1$       & $94.7$      & $92.2$       & $96.1$                  & $68.7$       & $90.4$       & $87.6$       & $92.5$       & $88.89$ \\
    OLoRA              & $\underline{90.9}$ & $95.0$      & $92.0$       & $96.3$                  & $69.0$       & $91.0$       & $87.9$       & $92.4$       & $89.32$ \\
    EVA                & $90.8$       & $95.0$      & $92.1$       & $96.2$                  & $69.5$       & $91.4$       & $88.8$       & $92.6$       & $89.55$ \\
    DoRA               & $89.5$       & $94.6$      & $89.9$       & $96.1$                  & $69.3$       & $91.0$       & $88.4$       & $92.4$       & $88.90$ \\
    \hline
    \midrule
    \mycc CNTLoRA-X-SVD      & \mycc $\textbf{91.1}$  & \mycc $\underline{96.1}$  & \mycc $\textbf{93.1}$  & \mycc $\underline{96.4}$  & \mycc  $69.7$       & \mycc  $91.6$       & \mycc  $\underline{89.1}$  & \mycc  $\textbf{93.2}$  &  \mycc $\textbf{90.03}$ \\
    \mycc  CNTLoRA-X-QR       &  \mycc $\underline{90.9}$  & \mycc  $\textbf{96.2}$  &  \mycc $\underline{92.9}$  & \mycc  $95.1$                  & \mycc  $69.5$       & \mycc  $91.2$       & \mycc  $88.9$       & \mycc  $92.1$       & \mycc  $89.6$ \\
     \mycc CNTLoRA-S-SVD      & \mycc  $90.2$       & \mycc  $95.9$      & \mycc  $92.2$       & \mycc  $96.1$                  & \mycc  $\textbf{70.1}$  & \mycc  $\underline{91.8}$  & \mycc  $88.2$       &  \mycc $\underline{92.8}$  & \mycc  $89.6$ \\
     \mycc CNTLoRA-S-QR       & \mycc  $90.8$       &  \mycc $95.7$      & \mycc  $92.8$       & \mycc  $\textbf{97.1}$  & \mycc  $69.6$       & \mycc  $91.6$       & \mycc  $88.8$       & \mycc  $92.6$       & \mycc  $\underline{89.8}$ \\
     \mycc CNTLoRA-Sh-SVD     &  \mycc $\underline{90.9}$  &  \mycc $95.3$      & \mycc  $92.3$       & \mycc  $\underline{96.4}$  &  \mycc $\underline{69.8}$  & \mycc  $91.3$       & \mycc  $\textbf{89.5}$  &  \mycc $92.4$       &  \mycc $89.7$ \\
     \mycc CNTLoRA-Sh-QR      &  \mycc $90.6$       & \mycc  $95.0$      & \mycc  $92.1$       & \mycc  $96.3$                  &  \mycc $69.5$       & \mycc  $\textbf{92.4}$  &  \mycc $\underline{89.1}$  &  \mycc $92.5$       & \mycc  $89.7$ \\
    \bottomrule
    \end{tabular}
}
\end{center}
\vskip -0.1in
\end{table*}

\subsection{Language Understanding}

\noindent \textbf{Quantitative results} We compare our methods against existing parameter-efficient fine-tuning approaches on the GLUE benchmark, which includes diverse language understanding tasks. Also, we compare our method combined with QR decomposition simultaneously. As shown in Table.~\ref{tab:glue_main}, most of CNTLoRA variants consistently achieve high performance. Especially, CNTLoRA-X-SVD consistently achieves the best or near-best performance in various datasets. These results highlight that CNTLoRA achieves strong performance not only when using SVD but also when initialized with QR decomposition, demonstrating the effectiveness of both approaches in enhancing fine-tuning.

\begin{figure*}[h]
    \centering
    \begin{subfigure}[b]{0.48\linewidth}
        \centering
        \includegraphics[width=1.0\linewidth]{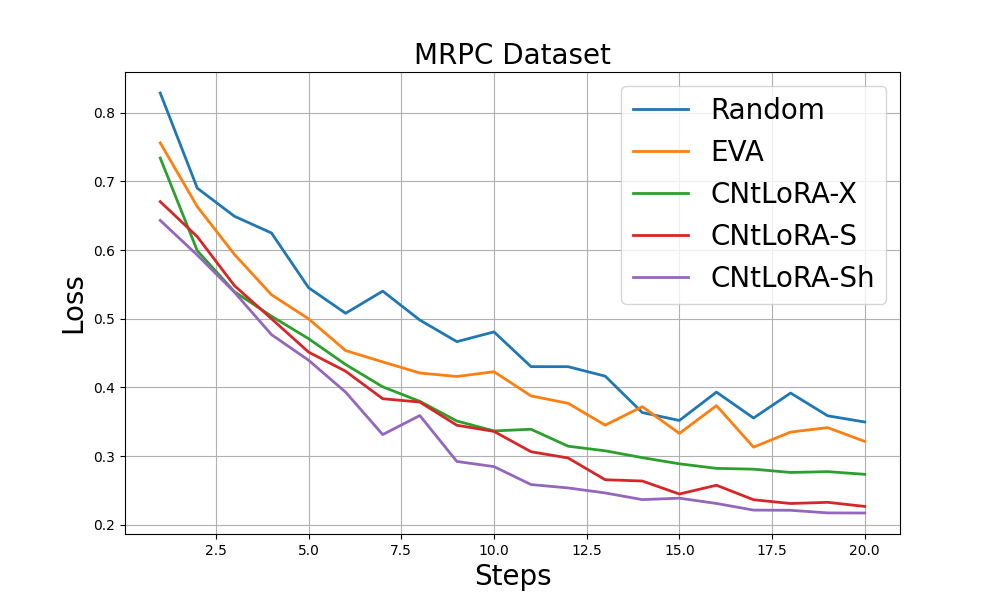}
        \vspace{-1.5em}
        \caption{Training loss for different initialization methods on the MRPC dataset.}
        \label{fig:text-loss-mrpc}
    \end{subfigure}
    \hfill
    \begin{subfigure}[b]{0.48\linewidth}
        \centering
        \includegraphics[width=1.0\linewidth]{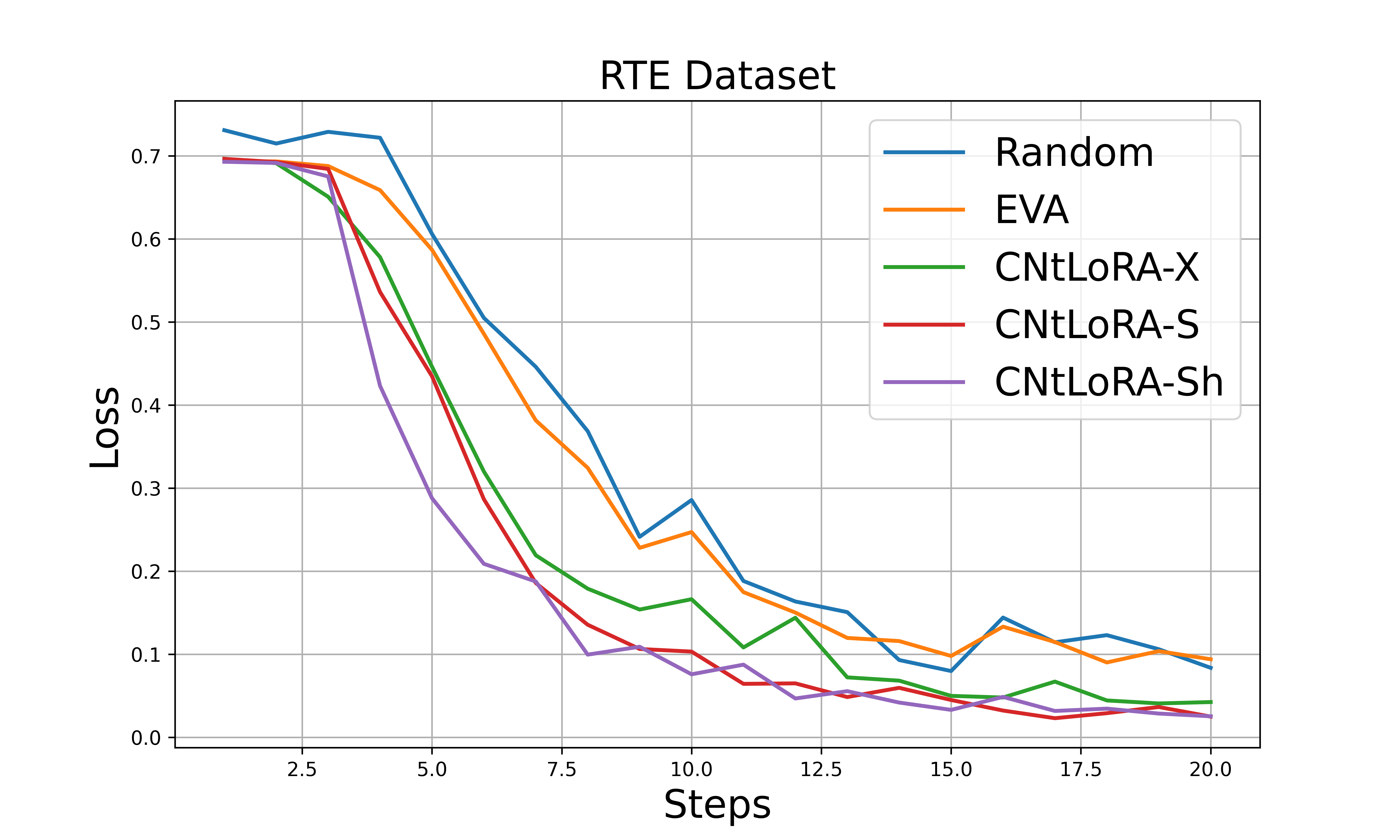}
        \vspace{-1.5em}
        \caption{Training loss for different initialization methods on the RTE dataset.}
        \label{fig:text-loss-rte}
    \end{subfigure}
    \caption{Plots showing how the training loss varies with different epochs for different initialization methods on the MRPC and RTE datasets.}
    \label{fig:combined-loss}
\end{figure*}

\noindent \textbf{Training curve}
We present the training curves for MRPC dataset and RTE dataset of the GLUE Benchmark in Figures~\ref{fig:text-loss-mrpc} and~\ref{fig:text-loss-rte}, respectively showing that our method achieves faster convergence compared to LoRA and EVA, particularly in the early epochs. We observe that our initialization better preserves pre-trained knowledge, allowing for a more stable adaptation to downstream tasks. We also find that our model maintains consistently lower training loss curves across multiple tasks, demonstrating improved fine-tuning stability.




\begin{figure*}[t]
    \centering
    \begin{subfigure}[b]{0.48\linewidth}
        \centering
        \includegraphics[width=1.0\linewidth]{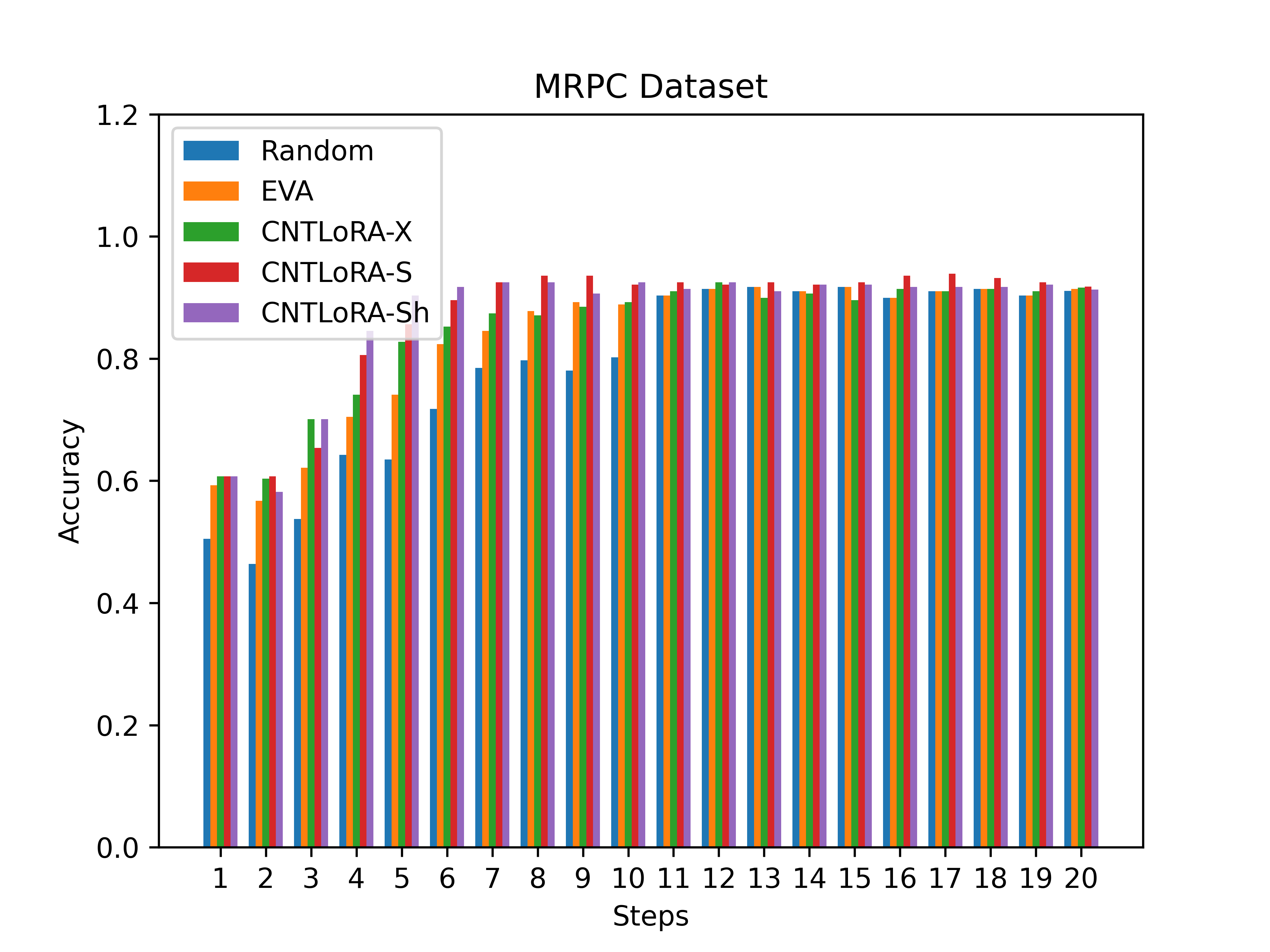}
        \vspace{-1.5em}
        \caption{Accuracy for different initialization methods on the MRPC dataset.}
        \label{fig:acc-mrpc}
    \end{subfigure}
    \hfill
    \begin{subfigure}[b]{0.48\linewidth}
        \centering
        \includegraphics[width=1.0\linewidth]{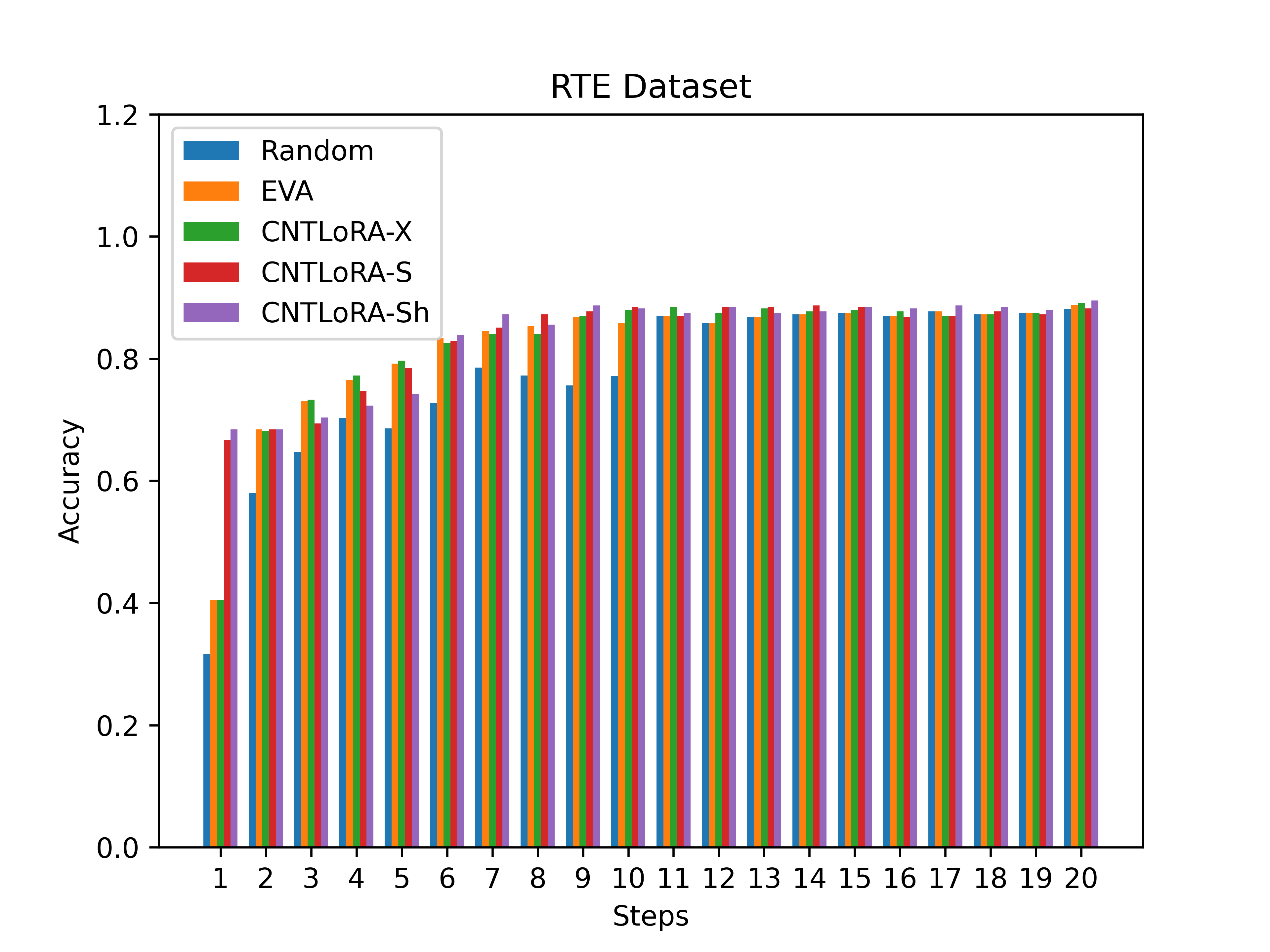}
        \vspace{-1.5em}
        \caption{Accuracy for different initialization methods on the RTE dataset.}
        \label{fig:acc-rte}
    \end{subfigure}
    \caption{Plots showing how the accuracy varies with different epochs for different initialization methods on the MRPC and RTE datasets.}
    \label{fig:combined-acc}
\end{figure*}

\noindent \textbf{Performance convergence}
We present the performance convergence results for MRPC and RTE of the GLUE Benchmark in Figures~\ref{fig:acc-mrpc} and~\ref{fig:acc-rte}, showing that our method achieves more stable performance compared to LoRA and EVA, with reduced fluctuations during fine-tuning. We observe that our initialization accelerates convergence, allowing the model to reach optimal performance with fewer fine-tuning steps. We also find that our approach consistently achieves higher final accuracy, demonstrating its effectiveness in enhancing fine-tuning efficiency and stability.

\end{document}